\documentclass[pdflatex,iicol,sn-basic,Numbered]{sn-jnl}

\usepackage{graphicx}
\usepackage{amsmath,amssymb}
\usepackage{booktabs}
\usepackage{multirow}

\begin{document}

% \title[Adaptive VIS--NIR Edge-Guided Multispectral Fusion]
% {Adaptive Visible--Near-Infrared Edge-Guided Multispectral Fusion for Historical Document Binarization}
\title[Cross-Spectral Feature Fusion for Multispectral Binarization] {Investigating Cross-Spectral Feature Fusion and Targeted Structural Guidance for Multispectral Document Binarization}

% \author*[1]{\fnm{Yasas} \sur{Savinda}}\email{e19356@eng.pdn.ac.lk}
% \author*[1]{\fnm{Maheshi Buddhinee} \sur{Dissanayake}}\email{maheshid@eng.pdn.ac.lk}
\author*{\fnm{Yasas} \sur{Savinda}}\email{e19356@eng.pdn.ac.lk}
\author*{\fnm{Maheshi Buddhinee} \sur{Dissanayake}}\email{maheshid@eng.pdn.ac.lk}

%\affil*[1]{%
\affil{%
\orgdiv{Department of Electrical and Electronic Engineering},
\orgname{University of Peradeniya},
\orgaddress{\city{Peradeniya}, \country{Sri Lanka}}}

% Add ORCID(s) when available using the syntax supported by the official template.
\abstract{%
\textbf{Purpose:} Historical document binarization remains challenging when faded ink, bleed-through, and background degradation exhibit wavelength-dependent behavior across visible and near-infrared observations. This study investigates whether adaptive cross-spectral feature fusion and targeted structural guidance can improve multispectral document binarization beyond fixed fusion and generic attention mechanisms.

\textbf{Methods:} Four registered SMADI bands at $500$, $600$, $700$, and $1000~\mathrm{nm}$ are processed using a multi-branch conditional generative adversarial network (cGAN). Visible, NIR, and structural feature streams are fused at encoder depth $L=3$ using spatial- and channel-wise softmax gating, while auxiliary edge maps derived from selected VIS--NIR band pairs provide targeted structural guidance. Nine controlled fusion and structural configurations are evaluated using five-fold document-level cross-validation, DIBCO-standard metrics, and Holm-corrected statistical testing.

\textbf{Results:} The $F3+\text{NIR}$ edge-guided configuration provides the strongest overall performance among the evaluated models, achieving mean $\mathrm{F_1}=0.7916$, $\mathrm{p\text{-}FM}=0.8044$, and $\mathrm{IoU}=0.6654$. Relative to the fixed Mid AVG baseline, it improves $\mathrm{F_1}$ by $0.0269$, $\mathrm{p\text{-}FM}$ by $0.0360$, and $\mathrm{IoU}$ by $0.0321$, with all three gains remaining statistically significant after Holm correction. Mean NRM and DRD are also reduced, although these reductions are not statistically significant after correction.

\textbf{Conclusion:} Targeted cross-spectral structural guidance provides an effective mechanism for regulating intermediate multispectral fusion, improving primary foreground-overlap performance while retaining the complete four-band observation.
}


\maketitle

% ============================================================
% 1 INTRODUCTION
% ============================================================
\section{Introduction}\label{sec:intro}

Historical documents are important linguistic, cultural, and historical sources, yet digital analysis is complicated by severe, spatially nonuniform degradation. Faded ink, bleed-through, stains, uneven illumination, aging, surface damage, and low foreground--background contrast can obscure strokes and reduce readability. Binarization therefore requires both background suppression and recovery of weak, thin, or discontinuous strokes without introducing false foreground structures. This has driven progression from global and adaptive thresholding toward supervised deep-learning and adversarial restoration methods \cite{DEGAN2022,GDB2024}, although severe degradation remains difficult when text--background separation is ambiguous in visible imagery.

Multispectral imaging (MSI) records the same document at multiple wavelengths, providing complementary evidence because ink, paper, stains, and degradation can respond differently across bands. Earlier work showed that hyperspectral and multispectral observations improve degraded-document visualization and restoration, with near-infrared (NIR) particularly useful when ink and substrate exhibit different spectral behavior \cite{Kim2011,Hedjam2013,Savinda2026MERCon}. This motivated multispectral text-extraction and binarization methods, including the ICDAR 2015 MS-TEx benchmark \cite{MSTEx2015}, target-detection methods \cite{MSIO2016}, CNN-based binarization \cite{Hollaus2019}, and MSdB-NMF \cite{MSdBNMF2020}, establishing that spectral diversity provides information unavailable from a single conventional image.

However, exploiting multiple spectral bands introduces a second problem: observations must be combined to preserve complementary information rather than treating all channels equally. Existing multispectral document methods have predominantly relied on spectral-band or subspace selection, analytical transformations, source decomposition, target detection, factorization, or direct learned mappings from spectral input to binary output \cite{MSIO2016,Hollaus2019,MSdBNMF2020}. These approaches demonstrate the value of multispectral information, but leave less explored how modality-specific visible and NIR features should be adaptively integrated within a deep restoration network. More generally, multimodal learning has shown that fusion location affects branch representations and that intermediate fusion preserves modality-specific processing before joint reasoning \cite{Baltrusaitis2019}.

Related developments in visible--infrared image fusion demonstrate the advantages of replacing fixed fusion rules with learned integration. Deep fusion models preserve complementary modality information, while adversarial formulations incorporate detail- and edge-sensitive objectives for structural preservation \cite{Ma2020}. These results are architecturally relevant, although generic visible--infrared fusion commonly considers thermal imagery, whereas historical-document MSI uses reflected NIR observations. In parallel, document-binarization research increasingly incorporates structural priors. For example, GDB combines Sobel-derived edge information with gated convolutions \cite{GDB2024}, while recent multispectral research shows that selected wavelength responses and ink-contour information can be highly discriminative for separating foreground from substrate \cite{KurarBarakat2026}. Collectively, these studies establish the value of learned gating, structural guidance, and wavelength-specific evidence, but largely investigate them separately.

Despite these advancements, existing literature remains bifurcated: generic vision frameworks investigate multimodal fusion strategies without considering document-specific stroke topology, whereas document restoration methods incorporate structural priors but operate largely on single-channel or RGB images. Crucially, no prior work has systematically investigated \emph{how}, \emph{where}, and \emph{under what structural guidance} visible and near-infrared (VIS--NIR) spectral bands should be integrated within a learned intermediate-fusion strategy for historical document restoration and downstream text legibility. 

This gap motivates the central objective of this work: to establish an adaptive, edge-guided multispectral fusion paradigm that bridges cross-spectral physics with structural text priors. Rather than forcing hard channel selection or unguided feature blending, we investigate whether a targeted VIS--NIR structural cue can dynamically regulate intermediate feature maps at encoder depth $L=3$, suppressing severe background artifacts such as stains, foxing, and bleed-through while recovering degraded or missing character-stroke geometry for binarisation and OCR readability.

This study substantially extends our preliminary multispectral-fusion work \cite{Savinda2026MERCon}, which compared early, intermediate, and late fusion and identified normalized additive fusion at encoder depth $L=3$ as the strongest baseline. That study established \emph{where} to fuse spectral information but not \emph{how} to adaptively control fusion. This journal extension replaces fixed averaging with learned modality and structural gating, adds targeted visible--NIR edge guidance and edge-aware supervision, shifts to full-page binarization, and expands evaluation to five-fold document-level cross-validation with statistical and computational profiling.

The principal contributions of this work are summarized as follows:
\begin{itemize}
    \item \textbf{Adaptive VIS--NIR Edge-Guided Fusion Architecture:} We propose a multi-branch framework for degraded document binarization that retains the full multispectral observation while employing an auxiliary visible--NIR structural representation to dynamically gate intermediate features at encoder depth $L=3$.
    
    \item \textbf{Controlled Architectural \& Spectral Source Ablation:} We conduct a systematic comparative study of nine fusion configurations, including fixed intermediate averaging, learned visible/NIR gating, edge-guided gating (with and without structural loss), generic spatial/channel attention (Squeeze-and-Excitation and CBAM), four-band structural guidance, and targeted spectral-source variants ($F_2$+NIR, $F_3$+NIR, and $F_4$+NIR). This isolates whether gains stem from adaptive fusion, generic attention, explicit structural conditioning, or specific spectral pairings.
    
    \item \textbf{Rigorous Page-Level Validation \& Statistical Profiling:} All architectures are evaluated using strict five-fold document-level cross-validation (ensuring zero patch leakage prior to extraction), validation-only threshold tuning, and full-page image reconstruction. Performance is assessed using standard DIBCO metrics ($F_1$, pseudo-$F_1$, IoU, NRM, and DRD), with pseudo-$F_1$ and DRD independently verified against official DIBCO reference tools, and backed by paired Wilcoxon signed-rank tests with Holm multiple-comparison correction.
    
    \item \textbf{Computational \& Reconstruction Profiling:} We characterize the proposed framework in terms of parameters, multiply--accumulate operations (MACs), floating-point operations (FLOPs), GPU memory footprint, patch-level inference latency, throughput, and overlap-sensitivity during full-page reconstruction.
\end{itemize}

Empirical evaluations confirm that structural guidance derived from the 600\,nm visible band and 1000\,nm NIR band ($F_3$+NIR) provides the strongest overall binarization performance among the evaluated configurations. Gains over the fixed intermediate averaging (Mid AVG) baseline are statistically significant after Holm correction for $F_1$, pseudo-$F_1$, and IoU ($p < 0.05$). These findings demonstrate that targeted cross-spectral structural guidance effectively regulates intermediate multispectral fusion without stroke fragmentation or spectral artifact transfer.

% Suggested flow:
% 1) degradation problem in historical documents
% 2) why multispectral/NIR evidence is useful
% 3) limits of current multispectral binarization methods
% 4) limits of generic adaptive fusion / edge-aware binarization
% 5) exact research gap
% 6) proposed idea
% 7) contributions
% 8) explicit relationship to the prior MERCon paper

% ============================================================
% 2 RELATED WORK
% ============================================================
% ============================================================
% 2 RELATED WORK
% ============================================================
\section{Related Work}\label{sec:related}

\subsection{Historical Document Binarization and Enhancement}\label{subsec:dib}

Document image binarization separates foreground text from background while preserving stroke geometry. Classical methods range from global Otsu thresholding \cite{Otsu1979} to locally adaptive Sauvola thresholding \cite{Sauvola2000}, which better handles varying illumination. These remain useful baselines, but combined faded ink, bleed-through, stains, texture variation, and localized degradation are difficult for a single handcrafted rule.

Deep-learning methods increasingly formulate binarization as supervised image-to-image prediction. Vo et al. \cite{Vo2018} introduced a hierarchical deeply supervised network in which multiple representation levels contribute to the binary output, combining coarse degradation suppression with fine boundary retention. Calvo-Zaragoza and Gallego \cite{Calvo2019} proposed a selectional auto-encoder that estimates foreground confidence directly, demonstrating end-to-end binarization across heterogeneous collections. DeepOtsu \cite{DeepOtsu2019} instead iteratively produces a cleaner representation for conventional thresholding.

Adversarial learning further extended this direction. Zhao et al. \cite{Zhao2019} used cascaded conditional GANs for degraded-document binarization, while DE-GAN \cite{DEGAN2022} supported binarization, deblurring, and watermark removal. Khamekhem Jemni et al. \cite{Jemni2022} added a recognition-oriented objective, showing that visually plausible restoration is insufficient if thin or degraded strokes are lost. These studies motivate reconstruction and adversarial objectives but do not explicitly exploit wavelength-dependent multispectral information.

Recent methods add stronger adaptive and structural mechanisms. GDB \cite{GDB2024} combines an Otsu-derived prior, Sobel gradients, an edge branch, and gated convolutions, while RA-GAN \cite{RAGAN2026} uses gradient-gated convolution, guided attention, and a gated feature pyramid. Both show that structural mechanisms improve binarization, but derive their priors from the same conventional input rather than complementary spectral observations.

\subsection{Multispectral Historical Document Analysis}\label{subsec:ms}

Multispectral and hyperspectral imaging provide registered observations at multiple wavelengths, allowing text, substrate, stains, and degradation to exhibit distinct responses. Kim et al. \cite{Kim2011} used hyperspectral information to enhance old documents, while Hedjam and Cheriet \cite{Hedjam2013} exploited visible and infrared behavior for restoration. Qureshi et al. \cite{Qureshi2019} further review hyperspectral imaging for historical information recovery, ink analysis, forensics, and cultural heritage.

The ICDAR 2015 MultiSpectral Text Extraction contest established a benchmark for multispectral text extraction \cite{MSTEx2015}. Early binarization methods relied on explicit spectral modeling. Farrahi Moghaddam and Cheriet \cite{Moghaddam2015} used multiple experts over different multispectral subspaces, showing that spectral combinations carry different discriminative information. MSIO \cite{MSIO2016} combined target detection with spatial segmentation, while Hollaus et al. \cite{HollausGMM2018} modeled foreground and background using Gaussian mixtures.

Later work adopted learned and factorized representations. Hollaus et al. \cite{Hollaus2019} applied a CNN directly to multispectral binarization, establishing a neural baseline. Rahiche et al. \cite{Rahiche2019} used graph-regularized non-negative matrix factorization for blind source separation, while MSdB-NMF \cite{MSdBNMF2020} reduced multispectral or hyperspectral observations before conventional binarization. These methods show that spectral dimensionality can be exploited without treating all channels identically.

%Kurar-Barakat and Dershowitz \cite{KurarBarakat2026} analyzed twelve multispectral observations of Dead Sea Scroll fragments and exploited discriminative spectral differences together with ink-contour information in an energy-minimization segmentation framework. This work is close to the present study in using selected spectral evidence and structural contours, but its processing remains based on spectral thresholding and graph-cut optimization. In contrast, the present work investigates whether a selected visible--NIR pair can provide an auxiliary structural representation that regulates learned intermediate feature fusion while the complete multispectral input remains available.
Recent work published in the International Journal on Document Analysis and Recognition (IJDAR) \cite{KurarBarakat2026} provides relevant evidence that selected wavelength responses and structural information can be more discriminative than indiscriminate spectral combination. The study analyzed twelve multispectral observations of Dead Sea Scroll fragments and exploited discriminative spectral differences together with ink-contour information in an energy-minimization segmentation framework.
This work is close to the present study in using selected spectral evidence and structural contours, but its processing remains based on spectral thresholding and graph-cut optimization. In contrast, the present work investigates whether a selected visible--NIR pair can provide an auxiliary structural representation that regulates learned intermediate feature fusion while the complete multispectral input remains available. Beyond pixel-level segmentation, such unified spectral-structural enhancement can benefit downstream optical character recognition (OCR) by removing severe background interference while reconstructing fragmented stroke geometry to improve machine text legibility.

\subsection{Adaptive Multimodal and Visible--Infrared Fusion}\label{subsec:fusion}

Multimodal representation learning broadly categorizes feature integration into early fusion at the input layer, intermediate fusion within latent feature spaces, and late fusion at the decision or output stage \cite{Baltrusaitis2019}. Intermediate fusion is advantageous when distinct modalities require domain-specific feature extraction before cross-modal interaction. For instance, FuseNet \cite{FuseNet2017} demonstrated the effectiveness of internally combining modality-specific feature streams within a convolutional architecture for RGB-D semantic segmentation.

To overcome static feature concatenation, adaptive paradigms dynamically weight multi-source contributions based on context. Gated Multimodal Units (GMUs) \cite{GMU2017} introduced multiplicative gating for context-dependent multimodal integration, providing a basis for replacing handcrafted fusion rules with learned representations. Attention mechanisms similarly offer spatial and channel-wise recalibration. Squeeze-and-Excitation (SE) networks \cite{SE2018} model inter-channel dependencies to emphasize informative feature maps, whereas CBAM \cite{CBAM2018} combines channel and spatial attention. While effective, these modules operate on domain-agnostic feature maps rather than document-specific stroke topographies, providing baselines for assessing explicit structural guidance.

Visible--infrared (VIS--IR) fusion provides another relevant precedent. DenseFuse \cite{DenseFuse2019} combines deep encoder features to preserve complementary visible and infrared information, while FusionGAN \cite{FusionGAN2019} uses adversarial learning to retain salient infrared targets and visible textures. Both demonstrate learned blending of complementary spectral inputs.

A physical distinction must be drawn between generic visible--infrared fusion and historical document analysis. Conventional VIS--IR fusion predominantly processes thermal infrared emissions (e.g., long-wave IR for heat signatures), whereas historical document multispectral imaging relies on reflected near-infrared (NIR, $\sim$1000\,nm) light. While deep adaptive fusion principles remain relevant, the physical basis of spectral complementarity differs: in document analysis, spectral contrast emerges from wavelength-dependent light absorption and scattering across ink pigments, aged substrates, stains, and bleed-through artifacts rather than thermal radiation.

\subsection{Structural Guidance for Fusion and Binarization}\label{subsec:structure}

Preserving edges and fine structures is a recurring concern in both multimodal image fusion and document binarization. In visible--infrared fusion, Ma et al. \cite{Ma2020} incorporated detail-preservation and target-edge-enhancement objectives into an adversarial framework to reduce boundary degradation and retain complementary source information. This shows that structural constraints can improve learned multimodal fusion, but the objective is generic fused-image generation rather than foreground/background separation in degraded historical documents.

In document binarization, GDB \cite{GDB2024} provides an algorithmic precedent by combining explicit Sobel edge maps with gated convolutions to maintain stroke integrity. Similarly, RA-GAN \cite{RAGAN2026} utilizes gradient-dependent gating and guided attention to handle spatially heterogeneous degradations. However, in both frameworks, the structural control signal is extracted solely from a single conventional document observation. These architectures therefore lack access to decoupled spectral channels where complementary, background-free structural cues can be isolated.

Conversely, multispectral document research confirms that specific spectral bands, inter-band differences, and ink-contour regions can be highly discriminative for foreground-background separation \cite{Moghaddam2015,KurarBarakat2026}. Nevertheless, existing techniques typically restrict these spectral cues to hard band selection, heuristic thresholding, matrix factorization, or localized graph-cut segmentation, rather than using them as dynamic control mechanisms within deep neural architectures.

Collectively, prior literature establishes four insights: (i)~multispectral observations reveal text and degradation cues inaccessible to standard RGB imaging; (ii)~intermediate learned fusion models cross-modal interactions; (iii)~gradient-aware mechanisms protect fine character-stroke topology during binarization; and (iv)~targeted spectral bands offer superior structural clarity. However, these paradigms have largely evolved in isolation. This work addresses the unresolved intersection by using a \emph{targeted cross-spectral structural representation} to govern intermediate multispectral feature fusion while preserving the full-spectrum observation.

% ============================================================
% 3 PROPOSED METHOD
% ============================================================

\section{Proposed Method}\label{sec:method}

\subsection{Problem Formulation}\label{subsec:problem}

Let a multispectral historical document page be represented by four spatially registered spectral observations,
\begin{equation}
\mathbf{X} = \left[X_{F2}, X_{F3}, X_{F4}, X_{F7}\right] \in \mathbb{R}^{H \times W \times 4},
\end{equation}
where $X_{F2}$, $X_{F3}$, and $X_{F4}$ denote visible wavelengths at 500, 600, and 700\,nm, respectively, and $X_{F7}$ corresponds to the 1000\,nm near-infrared (NIR) band. The input is partitioned into a visible stream $\mathbf{X}_{\mathrm{vis}} = [X_{F2}, X_{F3}, X_{F4}] \in \mathbb{R}^{H \times W \times 3}$ and an NIR stream $\mathbf{X}_{\mathrm{nir}} = X_{F7} \in \mathbb{R}^{H \times W \times 1}$.

The objective is to estimate a continuous document representation $\widehat{\mathbf{Y}} \in [-1, 1]^{H \times W}$ that separates foreground text from degraded background. A generator $G_{\theta}$ with learnable weights $\theta$ maps the input streams and auxiliary structural prior $\mathbf{E}$ to the output space,
\begin{equation}
\widehat{\mathbf{Y}} = G_{\theta}\left(\mathbf{X}_{\mathrm{vis}}, \mathbf{X}_{\mathrm{nir}}, \mathbf{E}\right).
\end{equation}
The auxiliary structural representation $\mathbf{E} = \mathcal{S}(X_{F3}, X_{F7})$ is constructed from a selected visible--NIR pair via a deterministic edge-extraction operator $\mathcal{S}(\cdot)$ (Section~\ref{subsec:preproc}). $\mathbf{E}$ provides targeted structural guidance without discarding or replacing any original spectral band in $\mathbf{X}$.

Rather than imposing fixed arithmetic rules, $G_{\theta}$ extracts modality-specific features and performs adaptive fusion at intermediate depth $L=3$. Let $\mathbf{H}_{\mathrm{vis}}^{(L)}$, $\mathbf{H}_{\mathrm{nir}}^{(L)}$, and $\mathbf{H}_{\mathrm{edge}}^{(L)}$ denote intermediate features from the visible, NIR, and structural branches, respectively. The fused representation is:
\begin{equation}
\mathbf{H}_{\mathrm{fuse}}^{(L)} = \Phi_{\theta}\left(\mathbf{H}_{\mathrm{vis}}^{(L)}, \mathbf{H}_{\mathrm{nir}}^{(L)}, \mathbf{H}_{\mathrm{edge}}^{(L)}\right),
\end{equation}
where $\Phi_{\theta}(\cdot)$ is a learnable spatial and channel-wise adaptive gating operator (Section~\ref{subsec:gate}).

Full-page predictions are converted to binary maps $\widehat{B}_{\tau}(p) \in \{0, 1\}$ using a global threshold $\tau$:
\begin{equation}
\widehat{B}_{\tau}(p) = 
\begin{cases}
0 & \text{if } \widehat{Y}(p) < \tau \quad (\text{foreground text}), \\
1 & \text{if } \widehat{Y}(p) \ge \tau \quad (\text{background}),
\end{cases}
\end{equation}
where $p$ denotes pixel location. Threshold $\tau$ is optimized on reconstructed validation pages and fixed during test evaluation.

\subsection{Multispectral Preprocessing and Structural Edge Construction}\label{subsec:preproc}

To prevent data leakage, document-level splitting precedes all statistical estimation and patch operations. Normalization parameters are derived only from training documents within each fold.

For each band $b \in \{F2, F3, F4, F7\}$, minimum and maximum intensities ($m_b, M_b$) are computed from the training set. Inputs are rescaled to $[-1, 1]$ via:
\begin{equation}
\widetilde{X}_{b} = 2 \left( \frac{X_b - m_b}{M_b - m_b + \epsilon} \right) - 1,
\end{equation}
where $\epsilon = 10^{-8}$ ensures numerical stability. 

Raw ground-truth maps, $Y_{\mathrm{raw}}$, are thresholded at 128 to enforce binary targets $Y(p) \in \{0, 255\}$ (0 for text, 255 for background). For training, $Y(p)$ is normalized to $\widetilde{Y} \in [-1, 1]$:
\begin{equation}
\widetilde{Y}(p) = 2 \left( \frac{Y(p)}{255} \right) - 1.
\end{equation}
The discrete $\{0, 255\}$ format is retained for padding and evaluation, while $\widetilde{Y}$ is used for loss calculation.

To prevent boundary artifacts during sliding-window inference, pages are symmetrically padded before patch extraction. Spectral inputs use reflection padding (\texttt{BORDER\_REFLECT\_101}), while target maps use a constant white value (255) to avoid artificial foreground edges.

The auxiliary structural representation $\mathbf{E}$ is generated at full-page resolution before patch extraction to avoid artificial boundary gradients. Construction comprises four steps:
\begin{enumerate}
\item \textbf{Intensity Normalization and Channel Averaging:}
The 600\,nm visible band ($X_{F3}$) and the 1000\,nm NIR band ($X_{F7}$) are used specifically to construct the auxiliary structural representation. These two bands are independently normalized to the 8-bit dynamic range ($[0,255]$) and combined using equal-weight averaging. Importantly, this spectral-pair selection applies only to the structural-map construction; the complete multispectral observation, comprising $X_{F2}$, $X_{F3}$, $X_{F4}$, and $X_{F7}$, remains available as input to the network:

\begin{equation}
\begin{aligned}
I_{\mathrm{VN}} = \frac{1}{2} \left[ 
  255 \cdot \frac{X_{F3} - \min(X_{F3})}
  {\max(X_{F3}) - \min(X_{F3}) + \epsilon} \right. \\
  \left. + \, 255 \cdot \frac{X_{F7} - \min(X_{F7})}
  {\max(X_{F7}) - \min(X_{F7}) + \epsilon} 
\right].
\end{aligned}
\end{equation}    
    \item \textbf{Local Contrast Enhancement:} Contrast-Limited Adaptive Histogram Equalization (CLAHE) is applied to $I_{\mathrm{VN}}$ with clip limit $c = 2.0$ over an $8 \times 8$ grid to enhance local stroke transitions:
    \begin{equation}
    I_{\mathrm{C}} = \operatorname{CLAHE}\left(I_{\mathrm{VN}}; \, c=2.0, \, g=8\times8\right).
    \end{equation}
    
    \item \textbf{Gradient Magnitude Extraction:} Horizontal ($G_x$) and vertical ($G_y$) gradients are computed using $3 \times 3$ Sobel operators, yielding magnitude $G$:
    \begin{equation}
    G = \sqrt{\left(S_x(I_{\mathrm{C}})\right)^2 + \left(S_y(I_{\mathrm{C}})\right)^2}.
    \end{equation}
    
    \item \textbf{Range Normalization:} $G$ is min--max rescaled to 8-bit space $E^{(8)}$ and mapped to $[-1, 1]$:
    % \begin{equation}
    % \widetilde{E} = 2 \left( \frac{E^{(8)}}{255} \right) - 1, \quad \text{where } E^{(8)} = 255 \cdot \frac{G - \min(G)}{\max(G) - \min(G) + \epsilon}.
    % \end{equation}
    \begin{equation}
    \begin{aligned}
    \widetilde{E} &= 2 \left( \frac{E^{(8)}}{255} \right) - 1, \\
    \text{where } E^{(8)} &= 255 \cdot \frac{G - \min(G)}{\max(G) - \min(G) + \epsilon}.
    \end{aligned}
    \end{equation} 
\end{enumerate}

Aligned spectral inputs $\mathbf{X}$, structural prior $\widetilde{E}$, and ground truth $\widetilde{Y}$ are extracted as $256 \times 256$ patches with 204-pixel stride ($\sim$20\% overlap), preserving registration across channels.

\subsection{Fixed Intermediate Fusion Baseline}\label{subsec:midavg}

The fixed intermediate-fusion architecture is the principal baseline for the proposed adaptive mechanism. It uses the intermediate scheme identified in our prior work \cite{Savinda2026MERCon}, where normalized additive fusion at encoder depth $L=3$ outperformed early- and late-fusion alternatives. The baseline is retrained and evaluated under the five-fold document-level cross-validation protocol in Section~\ref{subsec:cv}.

The multispectral input is partitioned into visible and near-infrared (NIR) streams,
\begin{equation}
\mathbf{X}_{\mathrm{vis}} = [X_{F2}, X_{F3}, X_{F4}], \qquad \mathbf{X}_{\mathrm{nir}} = [X_{F7}].
\end{equation}
The streams are encoded independently through the first three stages of the dual-branch network. Let $\mathbf{H}_{\mathrm{vis}}^{(3)}$ and $\mathbf{H}_{\mathrm{nir}}^{(3)}$ denote the representations at depth $L=3$:
\begin{equation}
\mathbf{H}_{\mathrm{vis}}^{(3)} = E_{\mathrm{vis}}^{(3)}\!\left(\mathbf{X}_{\mathrm{vis}}\right), \qquad \mathbf{H}_{\mathrm{nir}}^{(3)} = E_{\mathrm{nir}}^{(3)}\!\left(\mathbf{X}_{\mathrm{nir}}\right).
\end{equation}
Both feature maps have identical spatial and channel dimensions, permitting element-wise averaging:
\begin{equation}
\mathbf{H}_{\mathrm{avg}}^{(3)} = \frac{1}{2}\left(\mathbf{H}_{\mathrm{vis}}^{(3)} + \mathbf{H}_{\mathrm{nir}}^{(3)}\right).
\end{equation}
This fixes the spatial and channel-wise fusion weights to $\alpha_{\mathrm{vis}} = \alpha_{\mathrm{nir}} = 0.5$ across all samples and locations.

The unified representation $\mathbf{H}_{\mathrm{avg}}^{(3)}$ passes through the remaining shared encoder stages $E_{\mathrm{shared}}$ and U-Net decoder $D_{\theta}$ to produce $\widehat{\mathbf{Y}}_{\mathrm{avg}}$:
\begin{equation}
\widehat{\mathbf{Y}}_{\mathrm{avg}} = D_{\theta}\!\left(E_{\mathrm{shared}}\!\left(\mathbf{H}_{\mathrm{avg}}^{(3)}\right)\right).
\end{equation}

Because averaging has no learnable parameters, it assigns static weights regardless of local degradation or spectral contrast. This motivates the dynamic, content-adaptive architecture introduced next.

\subsection{Adaptive Visible--NIR Edge-Guided Fusion}\label{subsec:gate}

To overcome static averaging, the proposed generator replaces the two-stream baseline with learned three-stream dynamic gating and explicit structural guidance. The network accepts the complete multispectral tensor and auxiliary visible--NIR structural map $\widetilde{E}$ (Section~\ref{subsec:preproc}):
\begin{equation}
\mathbf{X}_{G} = \left[X_{F2}, X_{F3}, X_{F4}, X_{F7}, \widetilde{E}\right].
\end{equation}
These channels form three feature-extraction branches:
% \begin{equation}
% \mathbf{X}_{\mathrm{vis}} = [X_{F2}, X_{F3}, X_{F4}], \quad \mathbf{X}_{\mathrm{nir}} = [X_{F7}], \quad \text{and} \quad \mathbf{X}_{\mathrm{edge}} = [\widetilde{E}].
% \end{equation}
\begin{align}
\mathbf{X}_{\mathrm{vis}} &= [X_{F2}, X_{F3}, X_{F4}], \nonumber \\
\mathbf{X}_{\mathrm{nir}} &= [X_{F7}], \quad \text{and} \quad \mathbf{X}_{\mathrm{edge}} = [\widetilde{E}].
\end{align}

Each branch has three encoder stages, expanding channel capacity from 64 to 128 and 256. For a $256\times256$ patch, the representations at depth $L=3$ are:
\begin{align}
\mathbf{H}_{m}^{(3)} = E_{m}^{(3)}\!\left(\mathbf{X}_{m}\right) \in \mathbb{R}^{C \times H \times W}, \nonumber \\
\quad m \in \{\mathrm{vis}, \mathrm{nir}, \mathrm{edge}\},
\end{align}
% \begin{equation}
% \mathbf{H}_{m}^{(3)} = E_{m}^{(3)}\!\left(\mathbf{X}_{m}\right) \in \mathbb{R}^{C \times H \times W}, \quad m \in \{\mathrm{vis}, \mathrm{nir}, \mathrm{edge}\},
% \end{equation}
where $C=256$ and $H=W=32$.

To compute context-dependent fusion weights, the feature tensors are concatenated along the channel dimension:
\begin{equation}
\mathbf{H}_{\mathrm{cat}} = \operatorname{Concat}\!\left(\mathbf{H}_{\mathrm{vis}}^{(3)}, \mathbf{H}_{\mathrm{nir}}^{(3)}, \mathbf{H}_{\mathrm{edge}}^{(3)}\right) \in \mathbb{R}^{3C \times H \times W}.
\end{equation}
A $1\times1$ convolution maps the concatenated features to a logit tensor $\mathbf{Z} \in \mathbb{R}^{3C \times H \times W}$:
\begin{equation}
\mathbf{Z} = \operatorname{Conv}_{1\times1}\!\left(\mathbf{H}_{\mathrm{cat}}\right).
\end{equation}
Reshaping $\mathbf{Z}$ yields modality-specific logits $\left[\mathbf{Z}_{\mathrm{vis}}, \mathbf{Z}_{\mathrm{nir}}, \mathbf{Z}_{\mathrm{edge}}\right] \in \mathbb{R}^{C \times H \times W}$. Spatial- and channel-wise Softmax normalization is applied across $\mathcal{M} = \{\mathrm{vis}, \mathrm{nir}, \mathrm{edge}\}$:
\begin{equation}
A_m(c, p) = \frac{\exp\left(Z_m(c, p)\right)}{\displaystyle\sum_{k \in \mathcal{M}} \exp\left(Z_k(c, p)\right)}, \quad m \in \mathcal{M},
\end{equation}
where $c \in \{1, \dots, C\}$ denotes channel and $p \in \{1, \dots, H \times W\}$ spatial location. This guarantees unit-simplex normalization at each coordinate:
\begin{equation}
A_{\mathrm{vis}}(c, p) + A_{\mathrm{nir}}(c, p) + A_{\mathrm{edge}}(c, p) = 1.
\end{equation}

The fused representation $\mathbf{H}_{\mathrm{fuse}}^{(3)}$ is formed using element-wise Hadamard products ($\odot$):
\begin{equation}
\mathbf{H}_{\mathrm{fuse}}^{(3)} = \mathbf{A}_{\mathrm{vis}} \odot \mathbf{H}_{\mathrm{vis}}^{(3)} + \mathbf{A}_{\mathrm{nir}} \odot \mathbf{H}_{\mathrm{nir}}^{(3)} + \mathbf{A}_{\mathrm{edge}} \odot \mathbf{H}_{\mathrm{edge}}^{(3)}.
\end{equation}

After fusion at depth $L=3$, the streams merge and pass through the remaining shared encoder stages. A U-Net decoder with skip connections restores spatial resolution and produces the final single-channel output using a hyperbolic tangent ($\tanh$) activation:
\begin{equation}
\widehat{\mathbf{Y}} = G_{\theta}\!\left(\mathbf{X}_{\mathrm{vis}}, \mathbf{X}_{\mathrm{nir}}, \mathbf{X}_{\mathrm{edge}}\right).
\end{equation}

Unlike fixed averaging, this formulation allocates spatial and channel attention according to local signal quality, favoring visible features when strokes are clear, shifting toward NIR under severe visible-spectrum staining, or using $\mathbf{H}_{\mathrm{edge}}^{(3)}$ near degraded boundaries to preserve stroke topology.

% Visible branch: [F2,F3,F4]
% NIR branch: [F7]
% Edge branch: E
% Each branch: d1=64, d2=128, d3=256 channels
% Fusion after d3 (L=3)
%
% H_cat = concat(H_vis,H_nir,H_edge)
% 1x1 conv -> 3C logits
% reshape to [B,3,C,H,W]
% softmax over the 3 branch dimension
% H_f = A_vis*H_vis + A_nir*H_nir + A_edge*H_edge
%
% Emphasize that the gate is spatially and channel-wise adaptive.

\subsection{Conditional Adversarial Learning Objective}\label{subsec:cgan}

The framework is optimized as a supervised cGAN inspired by Pix2Pix \cite{Pix2Pix2017, Pix2pix2026}. Generator $G_{\theta}$ maps $\mathbf{X}_{G} = [X_{F2}, X_{F3}, X_{F4}, X_{F7}, \widetilde{E}]$ to a restored single-channel representation, while PatchGAN discriminator $D_{\phi}$ distinguishes generated outputs from paired ground truth conditioned on $\mathbf{X}_{G}$.

The conditional adversarial objective is:
\begin{equation}
\begin{aligned}
\mathcal{L}_{\mathrm{cGAN}}(G,D)
={}&
\mathbb{E}_{\mathbf{X}_{G},\mathbf{Y}}
\left[
\log D_{\phi}(\mathbf{X}_{G},\mathbf{Y})
\right]
\\
&\hspace{-1.8em}
+
\mathbb{E}_{\mathbf{X}_{G}}
\left[
\log\left(
1-D_{\phi}(\mathbf{X}_{G},G_{\theta}(\mathbf{X}_{G}))
\right)
\right].
\end{aligned}
\end{equation}

To enforce pixel-level alignment and preserve high-frequency stroke detail, the objective includes an $\ell_1$ reconstruction loss:
\begin{equation}
\mathcal{L}_{1}(G) = \mathbb{E}_{\mathbf{X}_{G}, \mathbf{Y}}\!\left[\left\| \mathbf{Y} - G_{\theta}(\mathbf{X}_{G}) \right\|_{1}\right].
\end{equation}
The $\ell_1$ norm provides a robust penalty that reduces blurring along thin text boundaries compared with squared error.

The total generator objective combines adversarial, pixel-level reconstruction, and edge-aware structural constraints:
\begin{equation}
\mathcal{L}_{G} = \mathcal{L}_{\mathrm{adv}} + \lambda_{1} \mathcal{L}_{1} + \lambda_{\mathrm{edge}} \mathcal{L}_{\mathrm{edge}},
\end{equation}
where $\lambda_{1} = 100$ and $\lambda_{\mathrm{edge}} = 5$ are fixed hyperparameters. 

Binary cross-entropy ($\mathrm{BCE}$) supervises adversarial updates. The discriminator loss $\mathcal{L}_{D}$ and generator adversarial term $\mathcal{L}_{\mathrm{adv}}$ are:

\begin{align}
\mathcal{L}_{D}
={}&
\frac{1}{2}\mathrm{BCE}\!\left(
D_{\phi}(\mathbf{X}_{G},\mathbf{Y}),
\mathbf{1}
\right)
\nonumber\\
&+
\frac{1}{2}\mathrm{BCE}\!\left(
D_{\phi}(\mathbf{X}_{G},\widehat{\mathbf{Y}}),
\mathbf{0}
\right),
\\
\mathcal{L}_{\mathrm{adv}}
={}&
\mathrm{BCE}\!\left(
D_{\phi}(\mathbf{X}_{G},\widehat{\mathbf{Y}}),
\mathbf{1}
\right).
\end{align}

where $\widehat{\mathbf{Y}} = G_{\theta}(\mathbf{X}_{G})$. Optimization alternates between updating $D_{\phi}$ with detached generated outputs and updating $G_{\theta}$ by end-to-end backpropagation.

% Pix2Pix/PatchGAN backbone.
%
% L_G = L_GAN + lambda_1 L_1 + lambda_edge L_edge
% lambda_1 = 100
% lambda_edge = 5
%
% IMPORTANT: lambda_edge=5 is a fixed design setting, not claimed as globally optimal.

\subsection{Edge-Aware Structural Loss}\label{subsec:edgeloss}

To penalize structural and boundary errors, an edge-consistency loss $\mathcal{L}_{\mathrm{edge}}$ is added to the objective. Tensors normalized to $[-1, 1]$ are rescaled to $[0, 1]$ via $\mathbf{Y}_{01} = \frac{1}{2}(\mathbf{Y} + 1)$ and $\widehat{\mathbf{Y}}_{01} = \frac{1}{2}(\widehat{\mathbf{Y}} + 1)$.

Gradient magnitudes $\mathcal{G}(\mathbf{I}) \in \mathbb{R}^{H \times W}$ are computed using Sobel operators $S_x$ and $S_y$:
\begin{equation}
\mathcal{G}(\mathbf{I}) = \sqrt{(S_x * \mathbf{I})^2 + (S_y * \mathbf{I})^2 + \epsilon},
\end{equation}
where $*$ denotes spatial convolution and $\epsilon = 10^{-8}$ ensures numerical stability during backpropagation.

With $\mathbf{G}_{\mathrm{pred}} = \mathcal{G}(\widehat{\mathbf{Y}}_{01})$ and $\mathbf{G}_{\mathrm{gt}} = \mathcal{G}(\mathbf{Y}_{01})$, the edge-aware loss measures mean absolute gradient difference:
\begin{equation}
\mathcal{L}_{\mathrm{edge}} = \mathbb{E}\!\left[\left\| \mathbf{G}_{\mathrm{pred}} - \mathbf{G}_{\mathrm{gt}} \right\|_{1}\right].
\end{equation}

While $\mathcal{L}_{1}$ enforces global intensity fidelity, $\mathcal{L}_{\mathrm{edge}}$ penalizes character erosion, stroke fragmentation, and boundary degradation. The complete generator loss is:
\begin{equation}
\mathcal{L}_{G} = \mathcal{L}_{\mathrm{adv}} + 100\mathcal{L}_{1} + 5\mathcal{L}_{\mathrm{edge}}.
\end{equation}

The reconstruction weight $\lambda_{1}=100$ follows Pix2Pix \cite{Pix2Pix2017} to retain pixel-level fidelity and baseline consistency. The fixed moderate weight $\lambda_{\mathrm{edge}}=5$ allows structural supervision to contribute without dominating adversarial and reconstruction objectives. 

Structural guidance acts in two complementary ways: (i) input-side channel $\widetilde{E}$ drives adaptive gating (Section~\ref{subsec:gate}), and (ii) output-side constraint $\mathcal{L}_{\mathrm{edge}}$ supervises boundary integrity.

 % Sobel gradient-magnitude consistency between prediction and GT.
 
\subsection{Full-Page Reconstruction and Inference}\label{subsec:recon}

To evaluate global topology and reduce patch-boundary artifacts, inference uses tiled full-page soft reconstruction. Padded documents are partitioned into aligned $256 \times 256$ patches with stride $s = 204$ pixels ($\sim 20\%$ overlap).

Let $\mathcal{P} = \{P_k\}_{k=1}^{K}$ denote the set of extracted patches for a document page. Each patch is processed independently by the generator to yield soft continuous predictions $\widehat{P}_k = G_{\theta}(P_k)$. Soft outputs are accumulated into a page-level intensity matrix $\mathbf{A}$ and a spatial coverage counter matrix $\mathbf{C}$:
\begin{equation}
\widehat{\mathbf{Y}}_{\mathrm{soft}}(p) = \frac{\mathbf{A}(p)}{\mathbf{C}(p)} = \frac{\sum_{k \in \mathcal{K}(p)} \widehat{P}_k(p)}{|\mathcal{K}(p)|},
\end{equation}
where $\mathcal{K}(p)$ denotes patches overlapping coordinate $p$.

Overlap averaging suppresses boundary seams and edge discontinuities. After cropping margins, the final binary map $\widehat{B}_{\tau}(p) \in \{0, 1\}$ is obtained by global thresholding:
\begin{equation}
\widehat{B}_{\tau}(p) = \begin{cases} 
0 (\text{foreground}), & \text{if } \widehat{\mathbf{Y}}_{\mathrm{soft}}(p) < \tau, \\
1 (\text{background}), & \text{if } \widehat{\mathbf{Y}}_{\mathrm{soft}}(p) \ge \tau.
\end{cases}
\end{equation}
The threshold $\tau$ is tuned only on reconstructed validation pages (Section~\ref{subsec:selection}) and applied unchanged to the test fold.
\subsection{Validation-Based Model and Threshold Selection}\label{subsec:selection}

To prevent data leakage, model and threshold selection use only the validation partition of each fold. Held-out test documents remain unseen during optimization and are evaluated once per fold.

Checkpoint screening uses validation epochs with soft outputs binarized at baseline threshold $\tau = 0.5$. Let $\overline{F}_{1}^{(e)}$ and $\overline{\mathrm{SSIM}}^{(e)}$ denote mean validation text $F_1$ and structural similarity at epoch $e$. Optimal parameters $\theta^{*} = \theta^{(e^{*})}$ are selected by lexicographic maximization:
\begin{equation}
e^{*} = \arg\max_{e} \left( \overline{F}_{1}^{(e)},\, \overline{\mathrm{SSIM}}^{(e)} \right),
\end{equation}
where $F_1$ is the primary criterion and SSIM resolves ties. 

Following checkpoint freeze at $G_{\theta^{*}}$, validation pages are reconstructed following Section~\ref{subsec:recon} and mapped to $[0,1]$ via $\widehat{Y}_{01} = \frac{1}{2}(\widehat{Y} + 1)$. The binarization threshold $\tau^{*}$ is optimized over a discretized candidate set $\mathcal{T} = \{0.10, 0.11, \dots, 0.90\}$ with a step size of $0.01$. Soft continuous predictions are thresholded to yield binary maps $\widehat{B}_{\tau}(p) \in \{0, 1\}$ ($0 = \text{foreground}$, $1 = \text{background}$):
\begin{equation}
\widehat{B}_{\tau}(p) = \begin{cases} 
0, & \text{if } \widehat{Y}_{01}(p) < \tau, \\
1, & \text{if } \widehat{Y}_{01}(p) \ge \tau.
\end{cases}
\end{equation}

Threshold optimization prioritizes boundary and stroke-topology preservation. Let $\overline{F}_{\mathrm{ps,val}}(\tau)$ and $\overline{F}_{1,\mathrm{val}}(\tau)$ denote mean validation pseudo-$F_1$ and text $F_1$ at threshold $\tau$. The optimal $\tau^{*}$ is selected by lexicographic maximization:
\begin{equation}
\tau^{*} = \arg\max_{\tau \in \mathcal{T}} \left( \overline{F}_{\mathrm{ps,val}}(\tau),\, \overline{F}_{1,\mathrm{val}}(\tau) \right).
\end{equation}
Here, pseudo-$F_1$ acts as the primary driver, with standard text $F_1$ serving as a secondary tie-breaker. (Note that this internal pseudo-$F_1$ screening metric is restricted strictly to parameter tuning; final reported p-FM metrics are computed independently via the standardized DIBCO evaluation toolkit described in Section~\ref{subsec:metrics}).

Once locked, $(G_{\theta^{*}}, \tau^{*})$ is applied directly to the unseen test set:
\begin{equation}
\widehat{B}_{\mathrm{test}} = \mathcal{B}_{\tau^{*}}\!\left(G_{\theta^{*}}(\mathbf{X}_{\mathrm{test}})\right).
\end{equation}
This two-stage validation is repeated independently across all five folds, so test metrics use model and threshold choices based only on training and validation data.
% ============================================================
% 4 EXPERIMENTAL PROTOCOL
% ============================================================

% % SMADI / MS-TEx lineage.
% % 30 usable paired documents.

\section{Experimental Protocol}\label{sec:protocol}

\subsection{Dataset and Spectral Configurations}\label{subsec:dataset}

Experiments are conducted on the Synchromedia Multispectral Ancient Document Images Dataset (SMADI), associated with the ICDAR 2015 MultiSpectral Text Extraction (MS-TEx) benchmark \cite{MSTEx2015}. The dataset provides spatially aligned multispectral observations of degraded historical pages with expert-annotated binary reference maps.

Following \cite{Savinda2026MERCon}, four registered bands are retained for 30 paired pages: visible channels at 500\,nm ($F2$), 600\,nm ($F3$), and 700\,nm ($F4$), plus a 1000\,nm NIR channel ($F7$). The observation vector is $\mathbf{X} = [X_{F2}, X_{F3}, X_{F4}, X_{F7}]$, partitioned into visible $\mathbf{X}_{\mathrm{vis}} = [X_{F2}, X_{F3}, X_{F4}]$ and NIR $\mathbf{X}_{\mathrm{nir}} = [X_{F7}]$ streams. The $F7$ channel is reflected NIR rather than thermal emission.

The four-band input $\mathbf{X}$ is retained across all edge-guided models. Band selection is restricted to constructing the auxiliary structural representation. In the primary architecture, the $F3$ (600\,nm) and $F7$ (1000\,nm) channels generate the auxiliary edge map, while the full input $\mathbf{X}$ remains accessible to the network encoder.

% % 18 train / 6 validation / 6 test documents per fold.
% % Every document is test once and validation once.
% % Split by document identifier before extracting patches.

\subsection{Five-Fold Document-Level Cross-Validation}\label{subsec:cv}

To prevent data leakage and evaluate full-page reconstruction, experiments use a strict five-fold cross-validation partitioned at the document level. The 30 SMADI pages are divided into five mutually exclusive 6-document folds: Fold~1 (\texttt{z92}, \texttt{z60}, \texttt{z802}, \texttt{z65}, \texttt{z38}, \texttt{z43}), Fold~2 (\texttt{z95}, \texttt{z82}, \texttt{z582}, \texttt{z100}, \texttt{z34}, \texttt{z64}), Fold~3 (\texttt{z35}, \texttt{z59}, \texttt{z58}, \texttt{z80}, \texttt{z27}, \texttt{z30}), Fold~4 (\texttt{z822}, \texttt{z31}, \texttt{z76}, \texttt{z90}, \texttt{z67}, \texttt{z97}), and Fold~5 (\texttt{z70}, \texttt{z37}, \texttt{z48}, \texttt{z592}, \texttt{z68}, \texttt{z36}).

For run $k \in \{1,\dots,5\}$, Fold $k$ is test ($N_{\mathrm{test}}=6$), the cyclic successor is validation ($N_{\mathrm{val}}=6$), and the remaining three folds are training ($N_{\mathrm{train}}=18$). The partitions are disjoint: $\mathcal{D}_{\mathrm{train}} \cap \mathcal{D}_{\mathrm{val}} = \mathcal{D}_{\mathrm{train}} \cap \mathcal{D}_{\mathrm{test}} = \mathcal{D}_{\mathrm{val}} \cap \mathcal{D}_{\mathrm{test}} = \varnothing$. Normalization and patch extraction occur only after assignment. Validation selects checkpoints and thresholds; test pages remain untouched until evaluation, yielding 30 independent reconstructed test pages per architecture.

\subsection{Compared Fusion Architectures}\label{subsec:comparators}

To isolate the effects of adaptive fusion, explicit edge conditioning, and spectral pairing, nine structural configurations are evaluated with identical training schedules, fusion depths ($L=3$), and validation pipelines (summarized in Table~\ref{tab:architectures}).

\begin{table*}[t]
\centering
\small
\caption{Comparative architectural study ($L=3$, evaluated under 5-fold document-level cross-validation).}
\label{tab:architectures}
\begin{tabular}{c l c c p{0.38\textwidth}}
\toprule
\textbf{No.} & \textbf{Configuration} & \textbf{Structural source} & $\boldsymbol{\lambda_{\mathrm{edge}}}$ & \textbf{Primary experimental objective} \\
\midrule
1 & Mid AVG & None & 0 & Fixed intermediate fusion baseline (equal-weight feature averaging). \\
2 & Visible/NIR gated fusion & None & 0 & Evaluates unguided, learned modality-dependent gating. \\
3 & Four-band edge-guided gate & $F2+F3+F4+F7$ & 0 & Tests structural input conditioning without edge-loss supervision. \\
4 & Four-band edge-guided gate & $F2+F3+F4+F7$ & 5 & Evaluates four-band structural guidance with explicit edge supervision. \\
5 & CBAM fusion & None & 0 & Baseline for domain-agnostic spatial and channel attention \cite{CBAM2018}. \\
6 & SE fusion & None & 0 & Baseline for domain-agnostic channel recalibration \cite{SE2018}. \\
7 & $F2+$NIR edge-guided gate & $F2+F7$ & 5 & Evaluates structural guidance derived from 500\,nm VIS + 1000\,nm NIR. \\
8 & $F3+$NIR edge-guided gate & $F3+F7$ & 5 & Primary proposed configuration (600\,nm VIS + 1000\,nm NIR). \\
9 & $F4+$NIR edge-guided gate & $F4+F7$ & 5 & Evaluates structural guidance derived from 700\,nm VIS + 1000\,nm NIR. \\
\bottomrule
\end{tabular}
\end{table*}

Configuration~1, \emph{Mid AVG}, uses fixed equal-weight fusion; Configuration~2 adds learned Visible/NIR gating without structural priors. Configurations~3--4 use four-channel edge maps to separate structural conditioning ($\lambda_{\mathrm{edge}}=0$) from edge supervision ($\lambda_{\mathrm{edge}}=5$). Configurations~5--6 use CBAM \cite{CBAM2018} and SE \cite{SE2018} as generic attention baselines. Configurations~7--9 keep identical topology ($\lambda_{\mathrm{edge}}=5$) while pairing 1000\,nm NIR with $F2$, $F3$, or $F4$.

% % 1 Mid AVG L3
% % 2 RGB/NIR gated L3
% % 3 Edge4Band gated L3, no edge loss
% % 4 Edge4Band gated L3, edge loss 5
% % 5 CBAM L3
% % 6 SE L3
% % 7 F2+NIR edge-guided L3, edge loss 5
% % 8 F3+NIR edge-guided L3, edge loss 5
% % 9 F4+NIR edge-guided L3, edge loss 5
% %
% % Call these "compared architectures/configurations", not nine proposed methods.

\subsection{Training Protocol}\label{subsec:training}

Networks are optimized independently in each fold using aligned $256\times256$ patches. Data augmentation (horizontal and vertical flips, $p=0.5$) is applied synchronously to input bands, structural maps, and ground-truth targets during training only.

Models are trained for 100 epochs per fold with the Adam optimizer ($\eta = 2\times10^{-4}$, $\beta_1=0.5$, $\beta_2=0.999$, batch size $B=8$, random seed 42). The generative objective is defined as:
\begin{equation}
\mathcal{L}_{G} = \mathcal{L}_{\mathrm{adv}} + 100\mathcal{L}_{1} + \lambda_{\mathrm{edge}}\mathcal{L}_{\mathrm{edge}},
\end{equation}
where $\lambda_{\mathrm{edge}}=5$ for edge-loss models and $0$ otherwise. Generator checkpoints are selected using validation text $F_1$ (using validation SSIM as a tie-breaker). Binarization thresholds are tuned only on validation pages reconstructed via overlapping patches (stride 204 pixels, $\sim$20\% overlap).

% % 100 epochs per fold
% % seed 42
% % Adam, learning rate 2e-4
% % beta1=0.5, beta2=0.999
% % batch size 8
% % synchronized horizontal/vertical flips during training only

\subsection{Evaluation Metrics}\label{subsec:metrics}

Evaluation uses full-page binarized outputs ($0=\text{foreground text}$, $1=\text{background}$) with five complementary metrics:
\begin{itemize}
    \item \textbf{Text $F_1$ Score:} Geometric precision-recall balance, $F_1 = \frac{2TP}{2TP + FP + FN}$.
    \item \textbf{Intersection over Union (IoU):} Direct spatial foreground overlap, $\mathrm{IoU} = \frac{TP}{TP + FP + FN}$.
    \item \textbf{Pseudo-FMeasure (p-FM):} Character-structure-aware metric prioritizing stroke boundary fidelity \cite{Ntirogiannis2013}. Computed with an official DIBCO-compatible reference pipeline.
    \item \textbf{Negative Rate Metric (NRM):} Balanced error metric, $\mathrm{NRM} = \frac{1}{2}\left(\frac{FN}{TP+FN} + \frac{FP}{FP+TN}\right)$. Lower is better.
    \item \textbf{Distance-Reciprocal Distortion (DRD):} Local structural distortion metric weighted by $8\times8$ ground-truth stroke neighborhoods \cite{DRD2004}, $\mathrm{DRD} = \frac{\sum_{k=1}^{S}\mathrm{DRD}_{k}}{N_{\mathrm{UBN}}}$. Computed with an official DIBCO reference tool. Lower is better.
\end{itemize}

% % Primary full-page binarization metrics:
% % F1, pseudo-FM, IoU, NRM, DRD.
% % Independently validate pseudo-FM and DRD using Doxa/BinBench.
% % Explicitly state foreground convention.

\subsection{Statistical Analysis}\label{subsec:stats}

Inferential statistics are calculated over the 30 paired full-page predictions. Non-parametric paired two-sided Wilcoxon signed-rank tests \cite{Demsar2006} compare each candidate architecture with the \emph{Mid AVG} baseline. 

Multiple testing is controlled using the Holm procedure \cite{Holm1979} at family-wise significance level $\alpha=0.05$, applied independently to the eight candidate comparisons within each metric family. Effect sizes are quantified by matched rank-biserial correlation, with document win/tie/loss counts and 10,000-resample bootstrap 95\% confidence intervals for the primary $F3+$NIR architecture.

\subsection{Classical Binarization Baselines}\label{subsec:classical}

Otsu \cite{Otsu1979} and Sauvola \cite{Sauvola2000} are evaluated as non-learning baselines on the 8-bit rescaled visible mean $I_{\mathrm{vis}} = \frac{1}{3}(X_{F2} + X_{F3} + X_{F4})$. Sauvola parameters $(w,k)$ are selected per fold using validation pseudo-$F_1$ over $w \in \{15, 25, 35, 51\}$ and $k \in \{0.1, 0.2, 0.3, 0.4\}$, yielding $(15,0.2)$, $(25,0.2)$, $(15,0.1)$, $(15,0.2)$, and $(15,0.2)$ for Folds 1--5. Classical results report $F_1$, IoU, and NRM for consistency with non-reference validated outputs.

% Otsu and Sauvola.

% ============================================================
% 5 RESULTS AND DISCUSSION
% ============================================================
\section{Results and Discussion}\label{sec:results}

\subsection{Overall Comparison of Fusion Architectures}\label{subsec:overall}

Table~\ref{tab:overall_results} summarizes full-page performance over 30 independently held-out pages across five folds. Values are mean $\pm$ standard deviation; p-FM and DRD use the independently validated DIBCO-compatible implementations in Section~\ref{subsec:metrics}. Higher F1, p-FM, and IoU and lower NRM and DRD are preferred.

\begin{table*}[t]
\centering
\caption{Full-page binarization performance of the nine fusion
architectures over 30 held-out SMADI documents. Values are reported as
mean $\pm$ standard deviation. Best values for each metric are shown in
bold.}
\label{tab:overall_results}

\setlength{\tabcolsep}{4pt}
\renewcommand{\arraystretch}{1.08}

\resizebox{\textwidth}{!}{%
\begin{tabular}{@{}lccccc@{}}
\toprule
\textbf{Architecture}
&
\textbf{F1 $\uparrow$}
&
\textbf{p-FM $\uparrow$}
&
\textbf{IoU $\uparrow$}
&
\textbf{NRM $\downarrow$}
&
\textbf{DRD $\downarrow$}
\\
\midrule

Mid AVG ($L=3$)
&
$0.7647 \pm 0.1232$
&
$0.7684 \pm 0.1134$
&
$0.6333 \pm 0.1486$
&
$0.1285 \pm 0.0871$
&
$5.9281 \pm 3.1901$
\\

Visible/NIR Gated ($L=3$)
&
$0.7651 \pm 0.1694$
&
$0.7663 \pm 0.1630$
&
$0.6414 \pm 0.1679$
&
$0.1162 \pm 0.1017$
&
$6.0284 \pm 3.7699$
\\

Edge4Band Gate ($\lambda_{\mathrm{edge}}=0$)
&
$0.7616 \pm 0.1416$
&
$0.7694 \pm 0.1315$
&
$0.6330 \pm 0.1641$
&
$0.1307 \pm 0.0946$
&
$5.9912 \pm 3.4993$
\\

Edge4Band Gate ($\lambda_{\mathrm{edge}}=5$)
&
$0.7853 \pm 0.1174$
&
$0.8021 \pm 0.1029$
&
$0.6600 \pm 0.1461$
&
$0.1227 \pm 0.0855$
&
$\mathbf{5.2537 \pm 2.7648}$
\\

CBAM Fusion ($L=3$)
&
$0.7502 \pm 0.1750$
&
$0.7557 \pm 0.1626$
&
$0.6247 \pm 0.1826$
&
$0.1318 \pm 0.1110$
&
$6.0324 \pm 3.3924$
\\

SE Fusion ($L=3$)
&
$0.7652 \pm 0.1647$
&
$0.7737 \pm 0.1535$
&
$0.6420 \pm 0.1751$
&
$0.1260 \pm 0.1029$
&
$5.8053 \pm 3.8226$
\\

$F2+$NIR Edge Gate ($\lambda_{\mathrm{edge}}=5$)
&
$0.7842 \pm 0.1090$
&
$0.7935 \pm 0.0963$
&
$0.6568 \pm 0.1360$
&
$\mathbf{0.1064 \pm 0.0704}$
&
$6.3810 \pm 5.2027$
\\

$F3+$NIR Edge Gate ($\lambda_{\mathrm{edge}}=5$)
&
$\mathbf{0.7916 \pm 0.0988}$
&
$\mathbf{0.8044 \pm 0.0862}$
&
$\mathbf{0.6654 \pm 0.1304}$
&
$0.1115 \pm 0.0697$
&
$5.7007 \pm 3.7990$
\\

$F4+$NIR Edge Gate ($\lambda_{\mathrm{edge}}=5$)
&
$0.7746 \pm 0.1371$
&
$0.7866 \pm 0.1172$
&
$0.6495 \pm 0.1614$
&
$0.1259 \pm 0.0967$
&
$5.4313 \pm 3.0338$
\\

\bottomrule
\end{tabular}%
}

\end{table*}

The $F3+$NIR edge-guided configuration gives the strongest overall performance, with mean F1=0.7916, p-FM=0.8044, and IoU=0.6654. Relative to Mid AVG, the absolute gains are 0.0269, 0.0360, and 0.0321, respectively; mean NRM decreases from 0.1285 to 0.1115 and DRD from 5.9281 to 5.7007.

The performance pattern nevertheless shows that no single architecture is
optimal for every error criterion. The $F2+$NIR configuration achieves the
lowest NRM of 0.1064, indicating the most favorable balance between
foreground and background classification errors according to this measure.
In contrast, the four-band edge-guided model with
$\lambda_{\mathrm{edge}}=5$ obtains the lowest DRD of 5.2537, suggesting
stronger suppression of locally weighted structural distortions. The
$F4+$NIR model also achieves a relatively low DRD of 5.4313 despite ranking
below the $F3+$NIR model in F1, p-FM, and IoU. These differences reinforce
the importance of assessing document binarization using complementary
structure-sensitive metrics rather than a single aggregate score.

A clear effect is also observed when the four-band structural-input
configuration is trained with and without the edge-aware loss. With
$\lambda_{\mathrm{edge}}=0$, the model obtains an F1 of 0.7616, p-FM of
0.7694, IoU of 0.6330, NRM of 0.1307, and DRD of 5.9912. Introducing the
edge-consistency term with $\lambda_{\mathrm{edge}}=5$ increases the
corresponding F1, p-FM, and IoU values to 0.7853, 0.8021, and 0.6600,
respectively, while reducing NRM to 0.1227 and DRD to 5.2537. Under the
controlled architecture and training protocol, these results indicate that
supplying structural information alone is less effective than jointly
incorporating structural guidance and output-side edge consistency.

Generic adaptive mechanisms provide limited gains. Visible/NIR gating gives F1=0.7651 versus 0.7647 for Mid AVG, although NRM falls to 0.1162. SE reaches F1/p-FM/IoU of 0.7652/0.7737/0.6420, while CBAM gives the lowest neural F1 (0.7502). Generic gating or attention therefore does not reproduce the gains from explicit cross-spectral structural guidance.

Among the three visible--NIR structural variants, the $F3+\text{NIR}$ configuration yields the highest $F_1$, $\text{p-FM}$, and $\text{IoU}$ scores, closely followed by $F2+\text{NIR}$. Because all variants share the same network backbone, fusion depth, loss weighting ($\lambda_{\mathrm{edge}}=5$), optimization protocol, and full multispectral input, these variations stem from the spectral pair used to construct the auxiliary structural map. This highlights the impact of targeted spectral guidance examined in Section~\ref{subsec:targeted}.

Overall, the gains arise from adaptive intermediate fusion coupled with an explicit visible--NIR structural representation, not merely learnable weights or generic attention. NRM and DRD differences also show that $F3+\text{NIR}$ is strongest for this dataset rather than universally optimal.

\subsection{Effect of Targeted Visible--NIR Structural Guidance}\label{subsec:targeted}

The overall comparison in Table~\ref{tab:overall_results} indicates that
the source used to construct the structural representation has a measurable
effect on binarization performance. To examine this behavior more directly,
Table~\ref{tab:edge_source_results} compares the four structurally guided
configurations trained with $\lambda_{\mathrm{edge}}=5$, together with the
four-band model trained without the edge-consistency loss.

\begin{table*}[t]
\centering
\small
\caption{Effect of structural-source selection and edge-aware supervision.
All models retain the complete $F2$--$F3$--$F4$--$F7$ multispectral input,
where $F2$, $F3$, and $F4$ denote the 500\,nm, 600\,nm, and 700\,nm visible bands, respectively, and $F7$ denotes the 1000\,nm near-infrared (NIR) band.
Only the source of the auxiliary structural representation and the use of the edge-consistency loss are varied.}
\label{tab:edge_source_results}
\begin{tabular}{lccccc}
\toprule
\textbf{Structural configuration}
&
\textbf{F1 $\uparrow$}
&
\textbf{p-FM $\uparrow$}
&
\textbf{IoU $\uparrow$}
&
\textbf{NRM $\downarrow$}
&
\textbf{DRD $\downarrow$}
\\
\midrule

Four-band, $\lambda_{\mathrm{edge}}=0$
&
0.7616
&
0.7694
&
0.6330
&
0.1307
&
5.9912
\\

Four-band, $\lambda_{\mathrm{edge}}=5$
&
0.7853
&
0.8021
&
0.6600
&
0.1227
&
\textbf{5.2537}
\\

$F2+F7$, $\lambda_{\mathrm{edge}}=5$
&
0.7842
&
0.7935
&
0.6568
&
\textbf{0.1064}
&
6.3810
\\

$F3+F7$, $\lambda_{\mathrm{edge}}=5$
&
\textbf{0.7916}
&
\textbf{0.8044}
&
\textbf{0.6654}
&
0.1115
&
5.7007
\\

$F4+F7$, $\lambda_{\mathrm{edge}}=5$
&
0.7746
&
0.7866
&
0.6495
&
0.1259
&
5.4313
\\

\bottomrule
\end{tabular}
\end{table*}

Among the three targeted visible--NIR configurations, the $F3+F7$
structural source provides the highest F1, p-FM, and IoU. Its mean F1 of
0.7916 exceeds the corresponding values of 0.7842 and 0.7746 obtained with
$F2+F7$ and $F4+F7$, respectively. Similar behavior is observed for p-FM,
where the $F3+F7$ model achieves 0.8044 compared with 0.7935 for $F2+F7$
and 0.7866 for $F4+F7$. The same ordering is retained for IoU, with values
of 0.6654, 0.6568, and 0.6495, respectively.

Because the three models use the same encoder depth, gating mechanism,
training objective, full multispectral input, and five-fold protocol, this
comparison isolates the wavelength pair used to construct the auxiliary
structural representation. Relative to $F2+F7$, the $F3+F7$ configuration
increases F1 by approximately 0.0074, p-FM by 0.0109, and IoU by 0.0086.
Relative to $F4+F7$, the corresponding improvements are approximately
0.0170, 0.0178, and 0.0158. These results suggest that the 600-nm and
1000-nm observations provide a particularly effective structural source for
recovering foreground stroke regions in the evaluated SMADI documents.

The $F3+F7$ advantage is not uniform across distortion measures.
The $F2+F7$ configuration obtains the lowest NRM, 0.1064, compared with
0.1115 for $F3+F7$. This indicates that the $F2+F7$ structural source
provides a slightly more favorable balance between foreground misses and
background false positives according to NRM. Conversely, $F4+F7$ obtains a
lower DRD of 5.4313 than the 5.7007 achieved by $F3+F7$, while the
four-band structural representation achieves the lowest overall DRD of
5.2537. Hence, the spectral pair producing the strongest foreground-overlap
measures is not necessarily the pair minimizing every form of local
structural distortion.

The four-band results provide another insight. When the
four-band structural map is supplied to the network without edge-aware
output supervision, the model achieves an F1 of 0.7616, p-FM of 0.7694,
IoU of 0.6330, NRM of 0.1307, and DRD of 5.9912. Retaining the same
structurally guided architecture while introducing
$\lambda_{\mathrm{edge}}=5$, increases F1 by approximately 0.0237, p-FM by
0.0327, and IoU by 0.0270. NRM decreases by approximately
0.0080 and DRD decreases by approximately 0.7375.

This contrast indicates that the auxiliary structural branch alone is not
sufficient to account for the performance obtained by the complete
edge-guided formulation. The combination of input-side structural
conditioning and output-side gradient consistency provides substantially
stronger results than the structurally guided architecture trained without
the edge-consistency term. Because only the fixed values
$\lambda_{\mathrm{edge}}=0$ and $\lambda_{\mathrm{edge}}=5$ are compared
in this experiment, the results should not be interpreted as establishing
5 as an optimal edge-loss weight.

The four-band and targeted-pair results further indicate that simply
aggregating all spectral observations when constructing the edge
representation is not always preferable. The four-band model produces the
lowest DRD, but the targeted $F3+F7$ representation achieves higher F1,
p-FM, and IoU. This supports the underlying hypothesis that the structural
cue used to regulate fusion can benefit from deliberate wavelength
selection rather than indiscriminate combination of all available bands.

Importantly, $F3+F7$ does not reduce the generator input to two bands. The full $\mathbf{X} = [X_{F2}, X_{F3}, X_{F4}, X_{F7}]$ remains available; $F3+F7$ only derives the auxiliary structural representation. The gains therefore reflect \emph{targeted structural guidance}, not exclusion of $F2$ or $F4$.

These findings identify the $F3+F7$ (600/1000\,nm) pair as the most effective structural source within our experimental setting. Although this does not imply universal optimality across degradation types, it shows that domain-targeted structural guidance outperforms generic attention and edge representations formed indiscriminately from all bands.

\subsection{Statistical Significance}\label{subsec:sig}

Document-level gains for $F3+\text{NIR}$ versus Mid AVG were tested with paired two-sided Wilcoxon signed-rank tests over all 30 held-out pages, with Holm adjustment across candidate architectures (Section~\ref{subsec:stats}). Table~\ref{tab:f3_statistics} reports statistics, win ratios, effect sizes ($r_{\mathrm{rb}}$), and adjusted significance.

\begin{table*}[t]
\centering
\small
\caption{Paired statistical comparison of the selected $F3+\text{NIR}$ edge-guided model against the Mid AVG baseline across 30 held-out test documents. Differences ($\Delta$) and rank-biserial correlations ($r_{\mathrm{rb}}$) are oriented such that positive values indicate improvement.}
\label{tab:f3_statistics}
\begin{tabular}{lcccccccc}
\toprule
\textbf{Metric} & \textbf{Mid AVG} & \textbf{$F3+$NIR} & \textbf{$\Delta$} & \textbf{Wins} & \textbf{$W$} & \textbf{$r_{\mathrm{rb}}$} & \textbf{$p$} & \textbf{$p_{\mathrm{Holm}}$} \\
\midrule
F1              & 0.7647           & 0.7916            & +0.0269           & 24/30         & 89           & 0.617                      & 0.00237      & \textbf{0.01894}              \\
p-FM            & 0.7684           & 0.8044            & +0.0360           & 23/30         & 94           & 0.596                      & 0.00348      & \textbf{0.02780}              \\
IoU             & 0.6333           & 0.6654            & +0.0321           & 24/30         & 89           & 0.617                      & 0.00237      & \textbf{0.01894}              \\
NRM             & 0.1285           & 0.1115            & +0.0170           & 21/30         & 122          & 0.475                      & 0.02210      & 0.15469                       \\
DRD             & 5.9281           & 5.7007            & +0.2274           & 20/30         & 132          & 0.432                      & 0.03842      & 0.29148                       \\
\bottomrule
\end{tabular}
\end{table*}

$F3+\text{NIR}$ significantly outperforms Mid AVG on all three primary overlap metrics after correction: $F_1$, $\text{p-FM}$, and $\text{IoU}$ have $p_{\mathrm{Holm}}<0.03$, rank-biserial effects $r_{\mathrm{rb}}\ge0.596$, and win ratios $\ge23/30$. Bootstrap 95\% confidence intervals remain positive for $F_1$ $[0.0014,0.0513]$, p-FM $[0.0080,0.0628]$, and IoU $[0.0009,0.0598]$.

$F3+\text{NIR}$ also reduces mean NRM from 0.1285 to 0.1115 and DRD from 5.9281 to 5.7007, but neither remains significant after Holm adjustment ($p_{\mathrm{Holm}}=0.15469$ and 0.29148). The DRD bootstrap interval $[-1.0420,1.3152]$ spans zero, so these reductions remain descriptive. Among other variants, only the four-band edge-guided model ($\lambda_{\mathrm{edge}}=5$) yields another Holm-significant gain over Mid AVG (p-FM: $W=98$, $r_{\mathrm{rb}}=0.578$, $p_{\mathrm{Holm}}=0.03265$).

In summary, non-parametric testing confirms that the advantages of targeted $F3+\text{NIR}$ structural guidance are consistent across pages and robust to multiple-comparison bias for spatial-overlap metrics, though non-significant for distance-based error criteria.
% Claim boundary:
% F3+NIR vs Mid AVG is significant after Holm for F1, p-FM, and IoU.
% Do NOT claim significant improvement for NRM or DRD.

\subsection{Comparison with Classical Binarization}\label{subsec:classicalres}
The multispectral framework is also evaluated against standard non-learning binarization algorithms: Otsu \cite{Otsu1979} and Sauvola \cite{Sauvola2000}. Both baselines operate on an 8-bit min--max rescaled visible-mean image:
\begin{equation}
I_{\mathrm{vis}} = \frac{1}{3}\left(X_{F2} + X_{F3} + X_{F4}\right).
\end{equation}
Otsu thresholding is applied globally without tuning. For Sauvola thresholding ($R=128$), fold-specific window sizes $w \in \{15, 25, 35, 51\}$ and sensitivity parameters $k \in \{0.1, 0.2, 0.3, 0.4\}$ are selected via validation pseudo-$F_1$ optimization (with smaller $w$, then smaller $k$ resolving ties). Table~\ref{tab:sauvola_settings} lists the fold-specific settings.

\begin{table}[t]
\centering
\small
\setlength{\tabcolsep}{4pt}
\renewcommand{\arraystretch}{1.05}
\caption{Fold-specific Sauvola parameters optimized strictly on validation partitions.}
\label{tab:sauvola_settings}
\begin{tabular*}{\columnwidth}{@{\extracolsep{\fill}}ccc@{}}
\toprule
\textbf{Fold} & \textbf{Window size ($w$)} & \textbf{$k$} \\
\midrule
1 & 15 & 0.2 \\
2 & 25 & 0.2 \\
3 & 15 & 0.1 \\
4 & 15 & 0.2 \\
5 & 15 & 0.2 \\
\bottomrule
\end{tabular*}
\end{table}

\begin{table}[t]
\centering
\small
\setlength{\tabcolsep}{4pt}
\renewcommand{\arraystretch}{1.05}
\caption{Performance comparison with classical binarization methods across 30 held-out test document pages.}
\label{tab:classical_results}
\begin{tabular*}{\columnwidth}{@{\extracolsep{\fill}}lccc@{}}
\toprule
\textbf{Method} &
\textbf{F1 $\uparrow$} &
\textbf{IoU $\uparrow$} &
\textbf{NRM $\downarrow$} \\
\midrule
Otsu & 0.5400 & 0.4076 & 0.1175 \\
Sauvola & 0.7436 & 0.6087 & 0.1178 \\
Mid AVG ($L=3$) & 0.7647 & 0.6333 & 0.1285 \\
\shortstack[l]{$F3+$NIR Edge Gate\\(Proposed)}
& \textbf{0.7916}
& \textbf{0.6654}
& \textbf{0.1115} \\
\bottomrule
\end{tabular*}
\end{table}

Table~\ref{tab:classical_results} reports mean performance over 30 held-out test pages. Global Otsu thresholding performs poorly ($F_1 = 0.5400$), especially when foreground strokes and background degradation overlap in the intensity histogram. Sauvola thresholding provides a stronger classical baseline ($F_1 = 0.7436$), confirming the value of spatially adaptive thresholding under non-uniform backgrounds.

The deep models nevertheless outperform the classical baselines. Fixed intermediate fusion (Mid AVG) improves upon Sauvola by $+0.0211$ $F_1$, showing that learned multispectral integration extracts information beyond local intensity thresholding. The proposed $F3+\text{NIR}$ edge-guided model achieves the highest performance ($F_1 = 0.7916$, $\mathrm{IoU} = 0.6654$, $\mathrm{NRM} = 0.1115$), with absolute gains of $+0.0480$ $F_1$ and $+0.0566$ $\mathrm{IoU}$ over Sauvola. Although Otsu achieves a seemingly competitive NRM ($0.1175$) due to metric balancing between false-negative and false-positive rates, its severe underperformance in $F_1$ and $\mathrm{IoU}$ confirms that comprehensive multi-metric evaluation is essential.

\subsection{Reconstruction-Overlap Sensitivity}\label{subsec:overlap}

% \begin{table}[t]
% \centering
% \small
% \setlength{\tabcolsep}{3pt} % Tightens column padding to fit 6 columns in \columnwidth
% \caption{Reconstruction-overlap sensitivity of the selected F3+NIR edge-guided model. No model retraining is performed.}
% \label{tab:overlap_results}
% \begin{tabular}{cccccc}
% \toprule
% \textbf{Overlap} & \textbf{F1 $\uparrow$} & \textbf{p-FM $\uparrow$} & \textbf{IoU $\uparrow$} & \textbf{NRM $\downarrow$} & \textbf{DRD $\downarrow$} \\
% \midrule
% 0\%  & 0.7617 & 0.7783 & 0.6275 & 0.1382 & 6.1510 \\
% 10\% & 0.7885 & 0.7995 & 0.6617 & 0.1141 & 5.7555 \\
% 20\% & 0.7916 & 0.8044 & 0.6654 & 0.1115 & 5.7007 \\
% 30\% & \textbf{0.7957} & \textbf{0.8053} & \textbf{0.6717} & \textbf{0.1078} & \textbf{5.5863} \\
% \bottomrule
% \end{tabular}
% \end{table}

\begin{table}[t]
\centering
\small
\setlength{\tabcolsep}{2pt}
\renewcommand{\arraystretch}{1.05}
\caption{Reconstruction-overlap sensitivity of the selected $F3+$NIR edge-guided model. No model retraining is performed.}
\label{tab:overlap_results}
\begin{tabular*}{\columnwidth}{@{\extracolsep{\fill}}cccccc@{}}
\toprule
\textbf{Overlap} &
\textbf{F1 $\uparrow$} &
\textbf{p-FM $\uparrow$} &
\textbf{IoU $\uparrow$} &
\textbf{NRM $\downarrow$} &
\textbf{DRD $\downarrow$} \\
\midrule
0\%  & 0.7617 & 0.7783 & 0.6275 & 0.1382 & 6.1510 \\
10\% & 0.7885 & 0.7995 & 0.6617 & 0.1141 & 5.7555 \\
20\% & 0.7916 & 0.8044 & 0.6654 & 0.1115 & 5.7007 \\
30\% & \textbf{0.7957} & \textbf{0.8053} & \textbf{0.6717} & \textbf{0.1078} & \textbf{5.5863} \\
\bottomrule
\end{tabular*}
\end{table}

A post hoc analysis evaluated the frozen $F3+\text{NIR}$ model at 0\% ($s=256$), 10\% ($s=230$), 20\% ($s=204$), and 30\% ($s=179$) overlap. For each stride, validation thresholding was re-optimized before locked test evaluation without retraining. Re-evaluating the 20\% baseline reproduced the original test binaries exactly. Table~\ref{tab:overlap_results} summarizes all 30 held-out pages.

The largest drop occurs at 0\% overlap ($F_1=0.7617$, DRD=6.1510), indicating severe boundary artifacts. A 10\% overlap restores $F_1$ to 0.7885, confirming the value of soft averaging near patch borders. Gains then diminish through 20\% and 30\%, with 30\% producing the best metrics ($F_1=0.7957$, IoU=0.6717, DRD=5.5863).

Holm-corrected paired Wilcoxon tests against the 20\% baseline show that 30\% overlap remains significantly better for $F_1$, $\mathrm{IoU}$, $\mathrm{NRM}$, and $\mathrm{DRD}$ ($p_{\mathrm{Holm}} < 0.05$). For a strict controlled comparison across architectures, the 20\% setting is retained for primary comparisons, balancing efficiency against incremental gains.

% Frozen selected model, no retraining:
% 0%, 10%, 20%, 30% overlap.
% Keep 20% as the fair setting for the nine-model comparison.

\subsection{Computational Complexity and Inference Efficiency}\label{subsec:efficiency}

The selected $F3+$NIR model was profiled without further optimization or compression; operation counts refer to one $256\times256$ patch and runtime to an NVIDIA Tesla T4 GPU.

The generator contains 56.31 million trainable parameters, requiring approximately 214.82~MiB for FP32 parameters; the saved checkpoint occupies 214.90~MiB. The footprint primarily reflects the three modality-specific encoder branches before $L=3$ fusion and the subsequent U-Net encoder--decoder.

A $256\times256$ patch requires approximately 20.68~GMACs, corresponding to 41.35~GFLOPs under the two-FLOPs-per-MAC convention. These supported-operation profiling estimates are approximate complexity indicators rather than hardware-independent runtime predictions.

Table~\ref{tab:complexity} summarizes the principal complexity and inference
measurements.

\begin{table}[t]
\centering
\small
\setlength{\tabcolsep}{4pt}
\renewcommand{\arraystretch}{1.05}
\caption{Computational characteristics of the selected $F3+$NIR
edge-guided generator. Runtime measurements were obtained on an NVIDIA
Tesla T4 GPU.}
\label{tab:complexity}

\begin{tabular*}{\columnwidth}{@{\extracolsep{\fill}}lr@{}}
\toprule
\textbf{Measure} & \textbf{Value} \\
\midrule
Trainable parameters & 56.31 M \\
FP32 parameter memory & 214.82 MiB \\
Checkpoint size & 214.90 MiB \\
Patch complexity (MACs) & 20.68 GMACs \\
Patch complexity (FLOPs) & 41.35 GFLOPs \\
Mean patch latency & 8.73 ms \\
Median patch latency & 8.68 ms \\
95th-percentile patch latency & 8.87 ms \\
Patch throughput & 114.51 patches/s \\
Peak GPU memory allocated & 460.58 MiB \\
Peak GPU memory reserved & 480.00 MiB \\
\bottomrule
\end{tabular*}

\end{table}

Single-patch inference is stable under the tested GPU conditions, yielding a mean latency of $8.73 \pm 0.57~\mathrm{ms}$ (median: $8.68~\mathrm{ms}$; $95^{\text{th}}$ percentile: $8.87~\mathrm{ms}$) with average throughput of approximately $114.5~\mathrm{patches/s}$. Peak allocated GPU memory during execution is $460.58~\mathrm{MiB}$ ($480~\mathrm{MiB}$ reserved), accounting for runtime tensors and network parameters.

For document $z92$, nine $256\times256$ patches at 20\% overlap required $0.0496 \pm 0.0056~\mathrm{s}$ model-only and $0.1056 \pm 0.0055~\mathrm{s}$ end-to-end. Overhead varies with page dimensions.

Although $30\%$ overlap yields peak binarization performance (Section~\ref{subsec:overlap}), the $20\%$ setting provides a favorable accuracy--cost trade-off. While practical for GPU deployment, the 56.31-million-parameter generator footprint motivates future work into model compression, knowledge distillation, and parameter-efficient fusion mechanisms.

% Report:
% parameter count
% MACs and FLOPs with explicit convention
% T4 patch latency, throughput, peak GPU memory
% representative full-page time
% State clearly that timing is hardware dependent.

\subsection{Qualitative Analysis}
\label{subsec:qualitative}

\begin{figure*}[t]
\centering
\setlength{\tabcolsep}{3pt}
\renewcommand{\arraystretch}{1.05}

\begin{tabular}{lccc}
\toprule
\textbf{Method} &
\textbf{$z27$} &
\textbf{$z35$} &
\textbf{$z80$} \\
\midrule

Mid AVG $L=3$ &
\includegraphics[height=2.40cm]{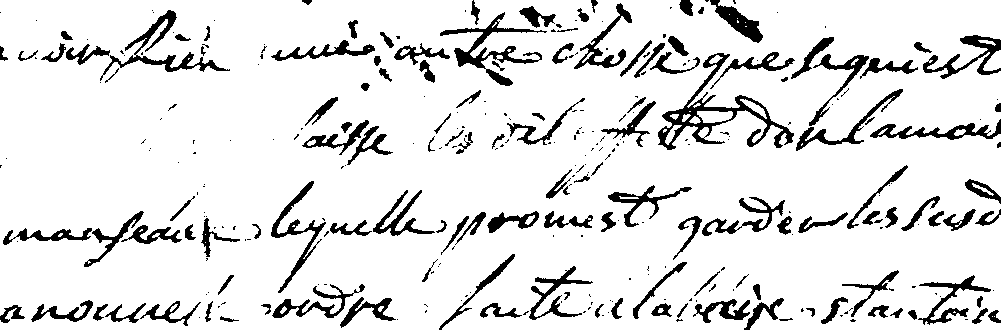} &
\includegraphics[height=2.40cm]{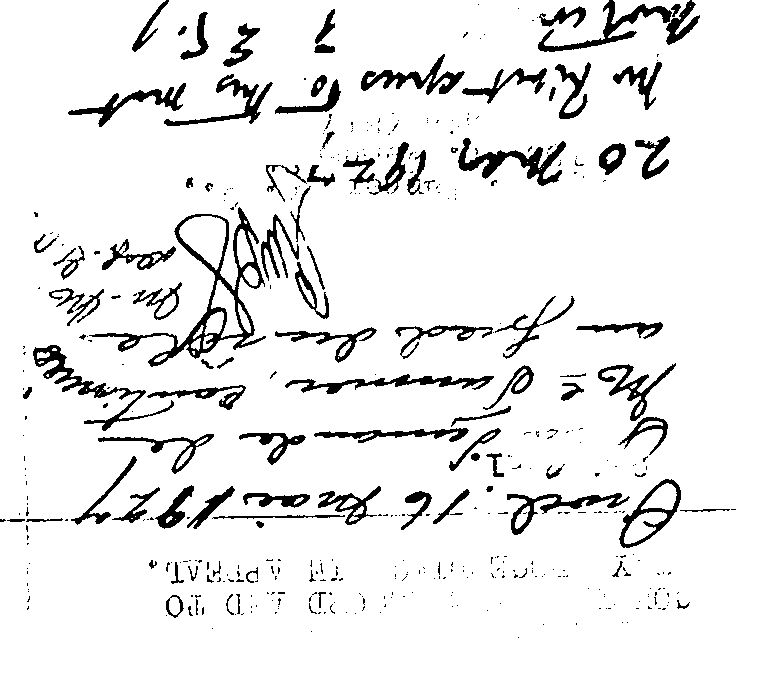} &
\includegraphics[height=2.40cm]{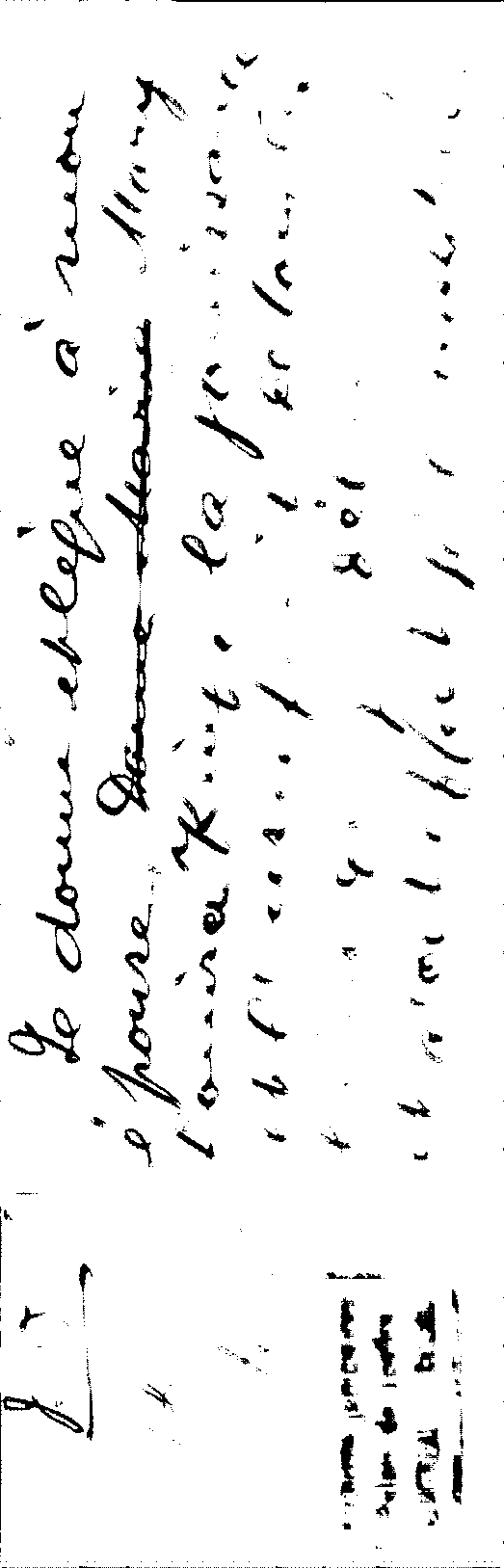}
\\

Visible/NIR Gated &
\includegraphics[height=2.40cm]{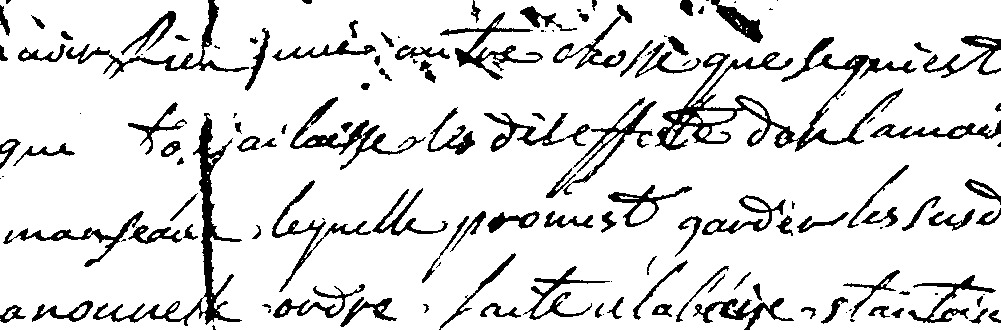} &
\includegraphics[height=2.40cm]{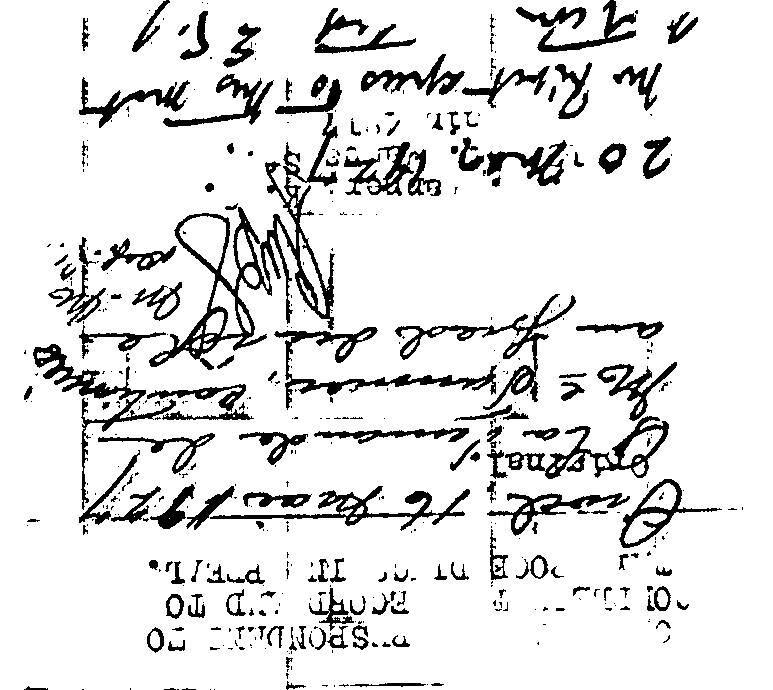} &
\includegraphics[height=2.40cm]{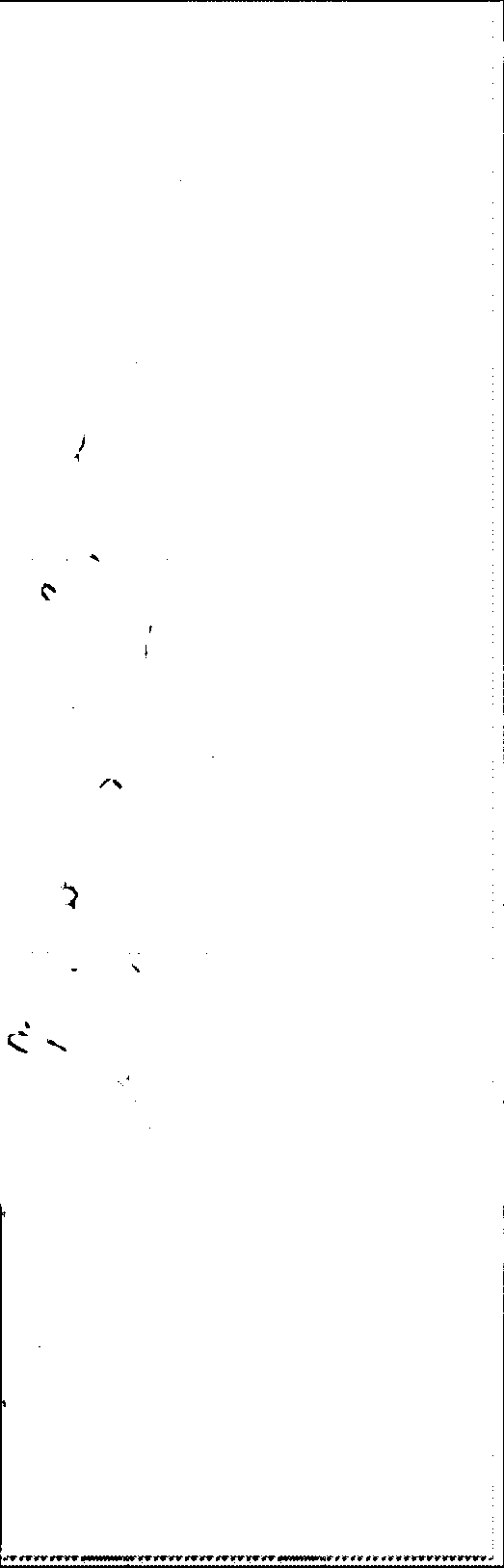}
\\

Edge4Band, $\lambda_{\text{edge}}=0$ &
\includegraphics[height=2.40cm]{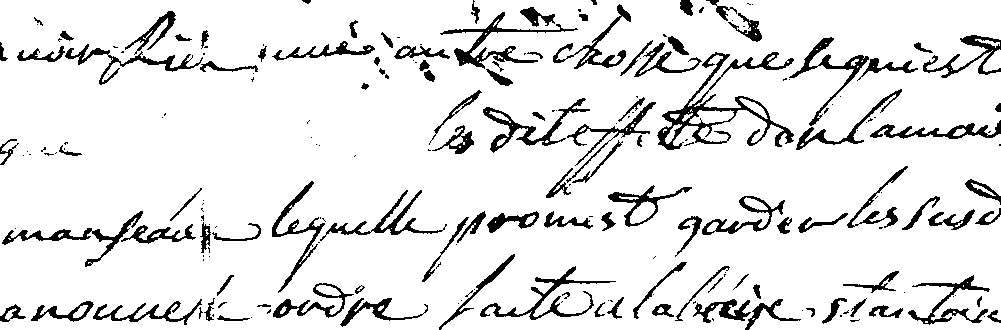} &
\includegraphics[height=2.40cm]{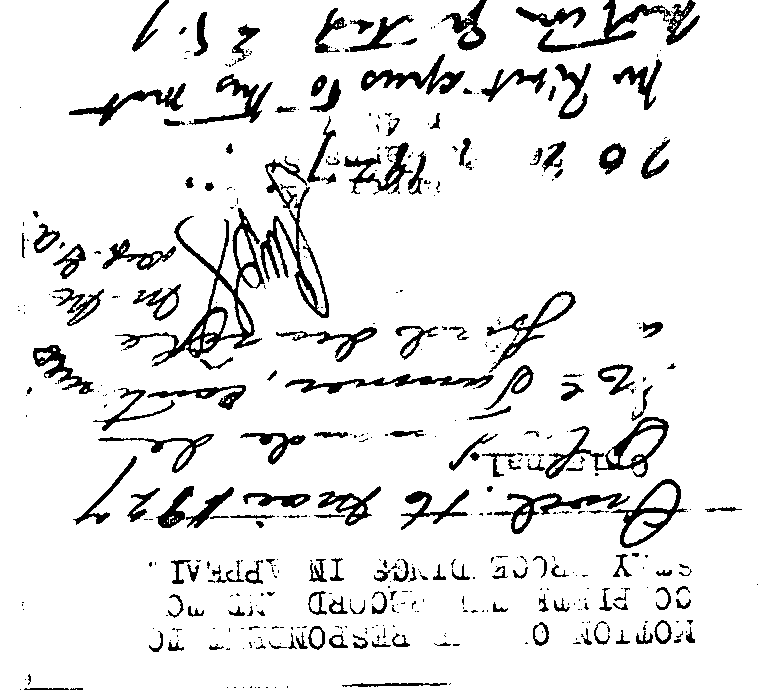} &
\includegraphics[height=2.40cm]{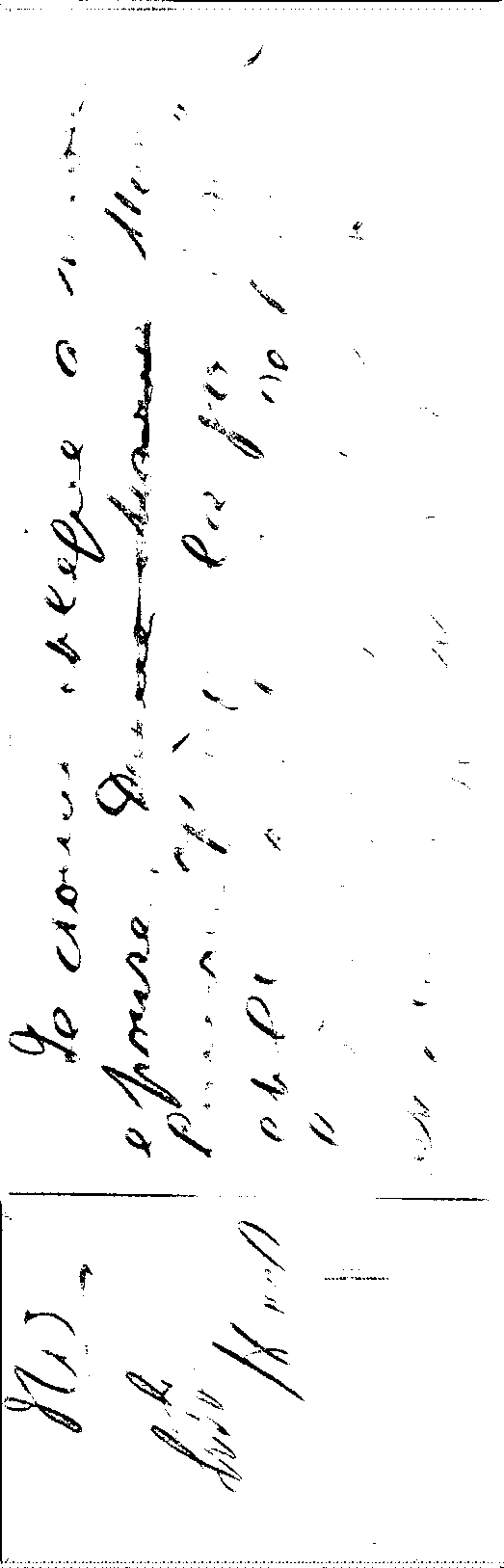}
\\

Edge4Band, $\lambda_{\text{edge}}=5$ &
\includegraphics[height=2.40cm]{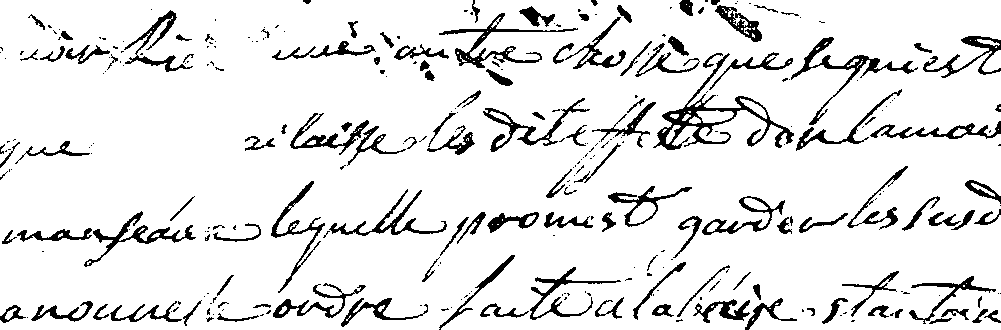} &
\includegraphics[height=2.40cm]{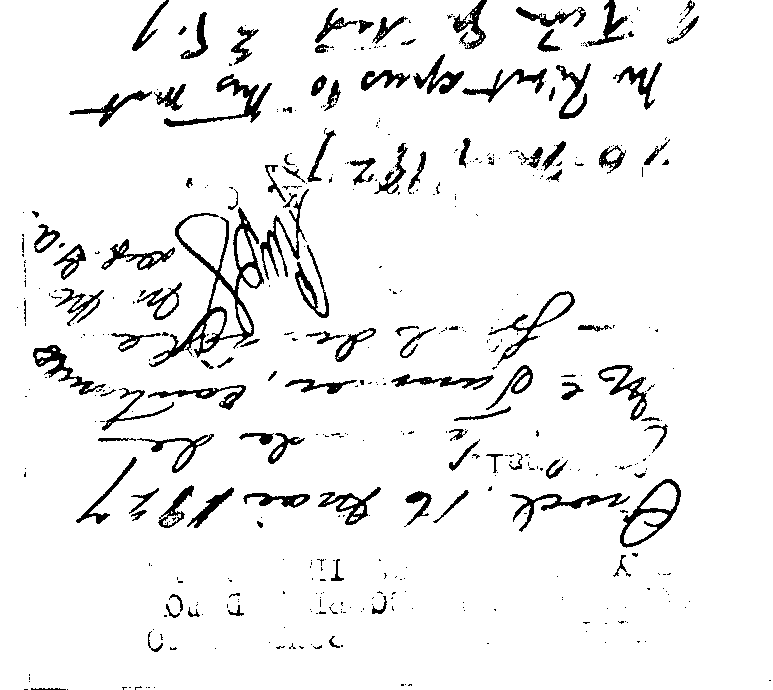} &
\includegraphics[height=2.40cm]{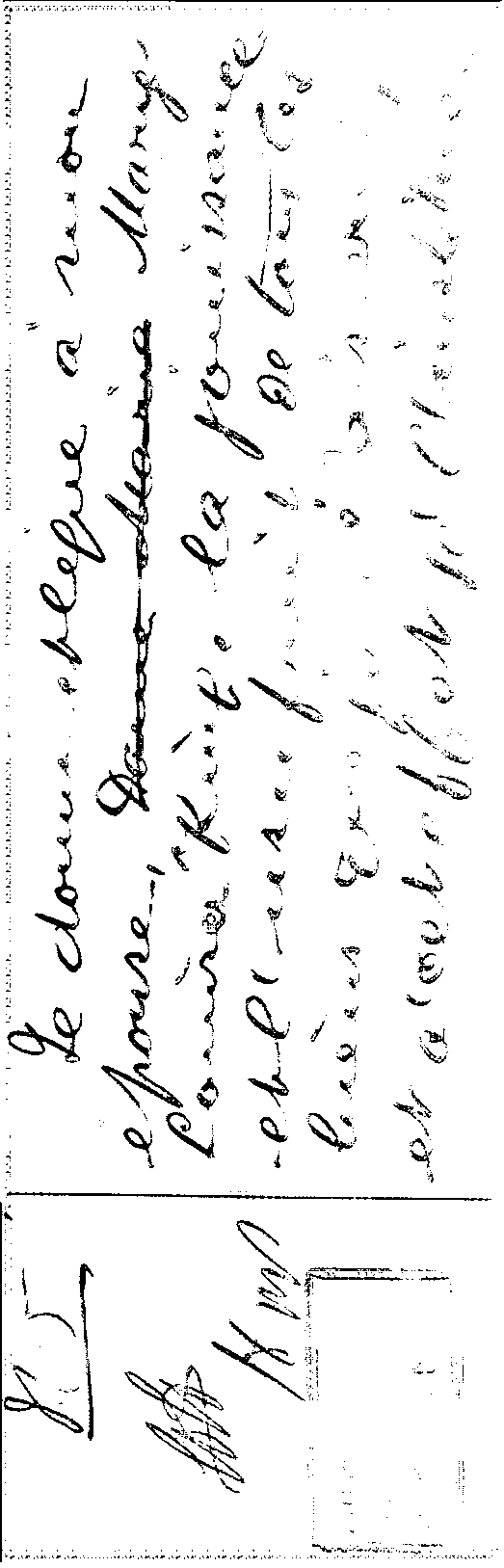}
\\

CBAM Fusion &
\includegraphics[height=2.40cm]{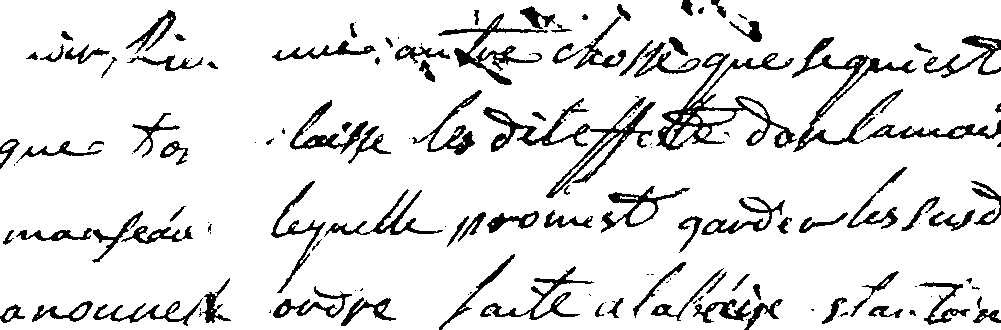} &
\includegraphics[height=2.40cm]{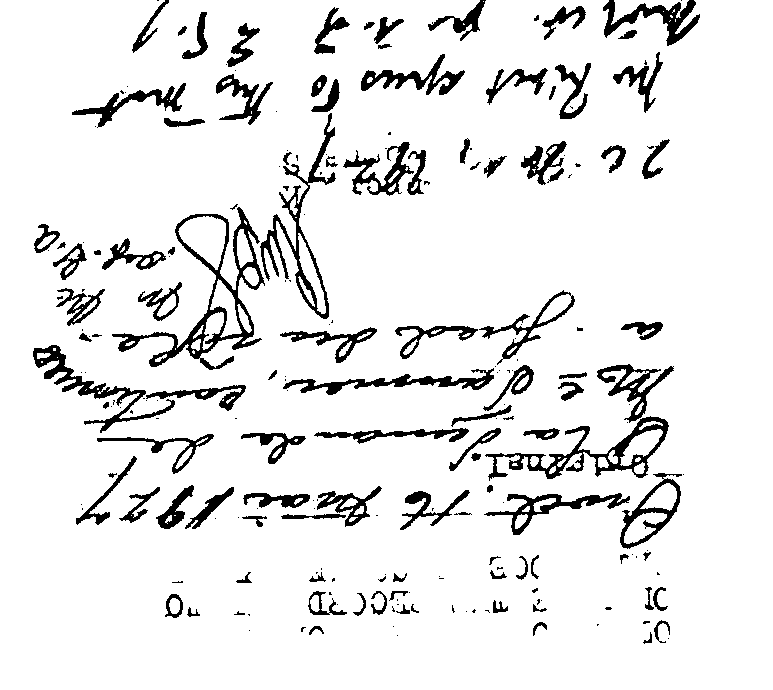} &
\includegraphics[height=2.40cm]{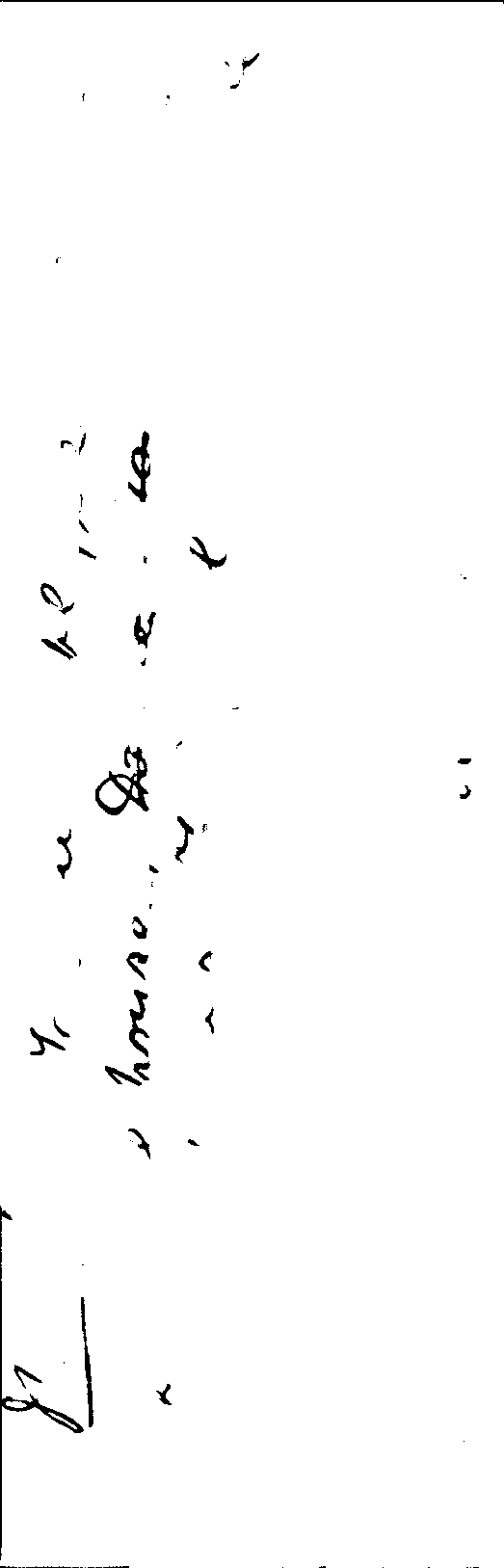}
\\

\bottomrule
\end{tabular}

\caption{Qualitative comparison of baseline, gated, and channel-attention architectures on held-out documents ($z27$, $z35$, and $z80$), using validation-selected thresholds.}
\label{fig:qualitative_a}
\end{figure*}

\begin{figure*}[t]
\centering
\setlength{\tabcolsep}{3pt}
\renewcommand{\arraystretch}{1.05}

\begin{tabular}{lccc}
\toprule
\textbf{Method} &
\textbf{$z27$} &
\textbf{$z35$} &
\textbf{$z80$} \\
\midrule

SE Fusion &
\includegraphics[height=2.40cm]{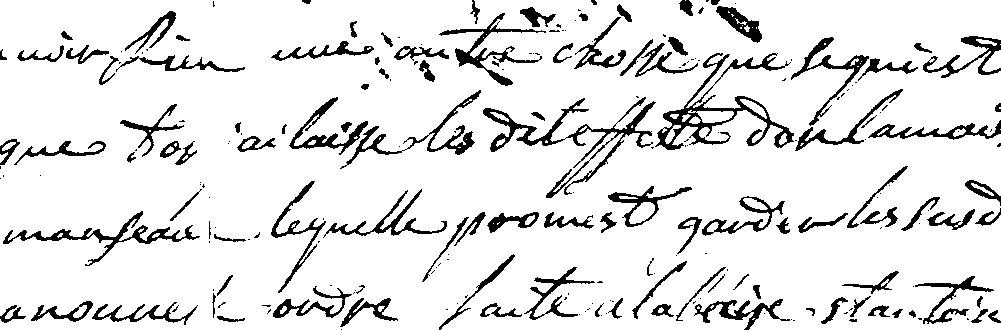} &
\includegraphics[height=2.40cm]{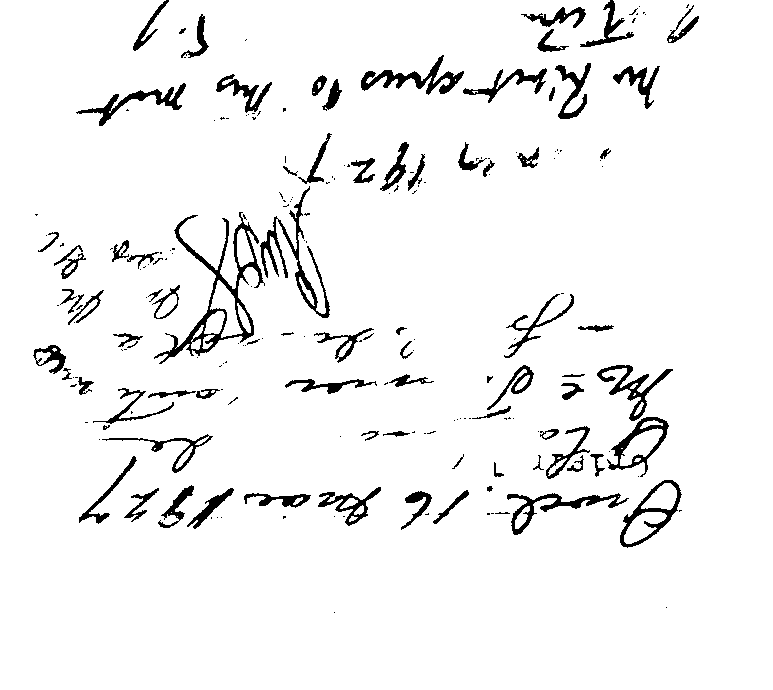} &
\includegraphics[height=2.40cm]{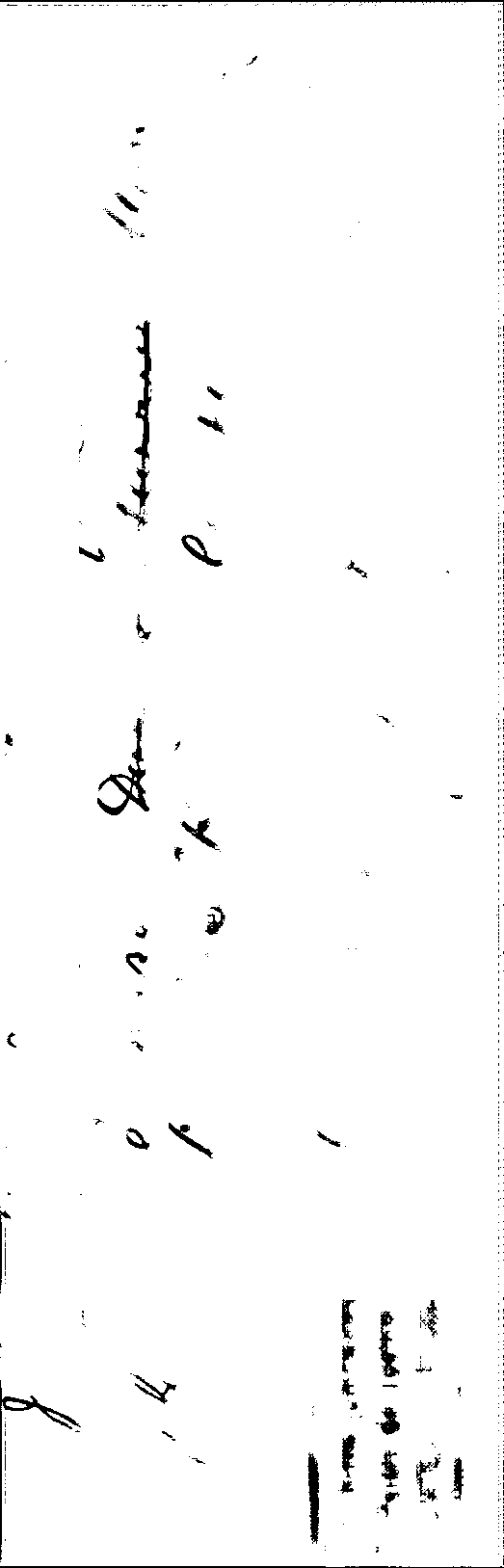}
\\

$F2+$NIR Edge-Guided &
\includegraphics[height=2.40cm]{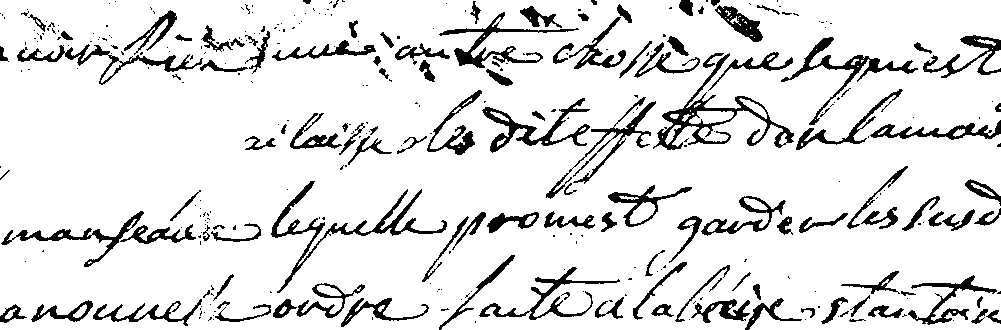} &
\includegraphics[height=2.40cm]{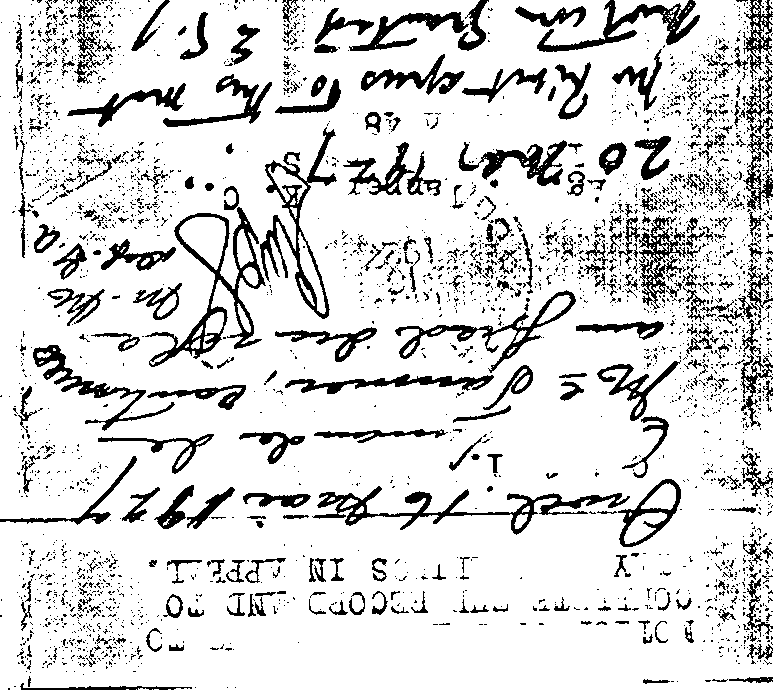} &
\includegraphics[height=2.40cm]{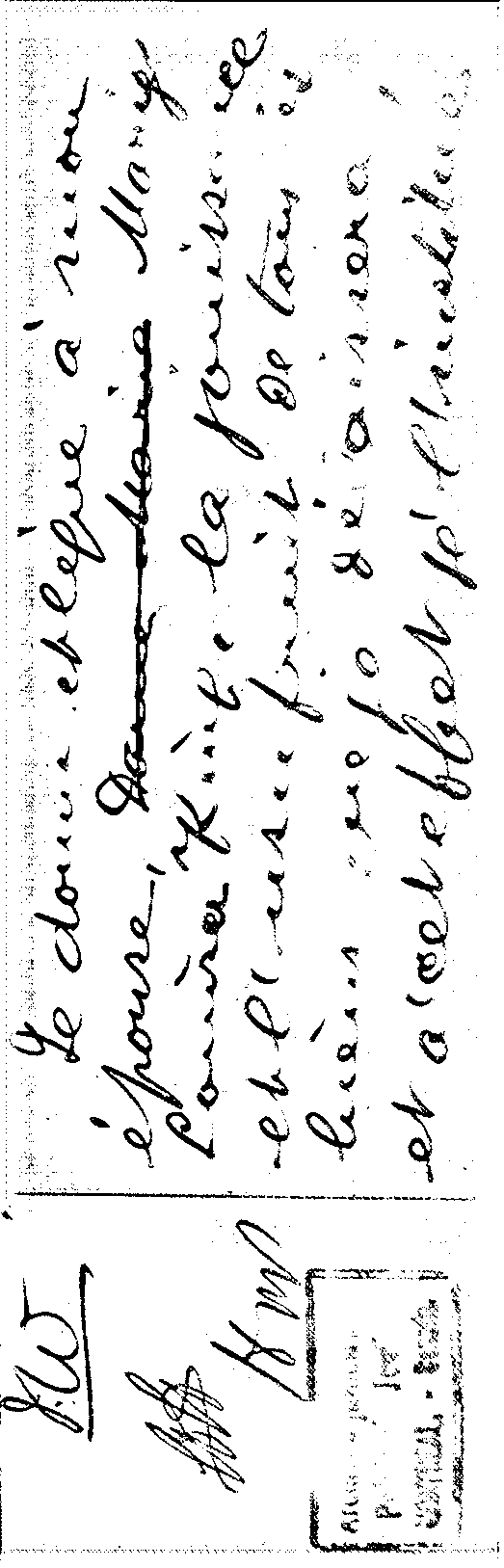}
\\

%\textbf{$F3+$NIR Edge-Guided (Proposed)} &
\shortstack[l]{\textbf{$F3+$NIR Edge-Guided}\\\textbf{(Proposed)}} &
\includegraphics[height=2.40cm]{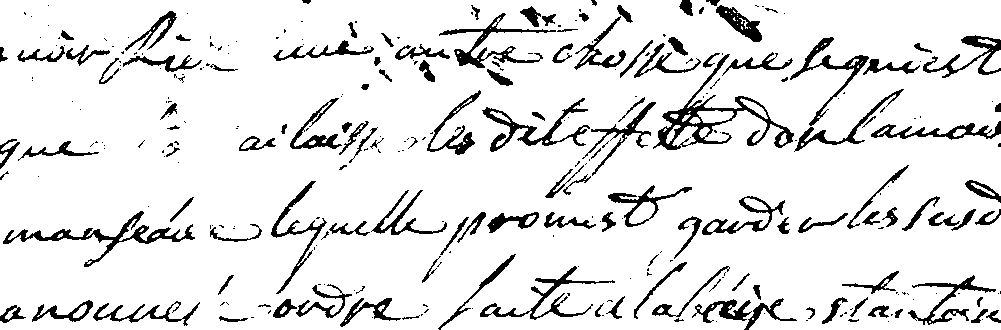} &
\includegraphics[height=2.40cm]{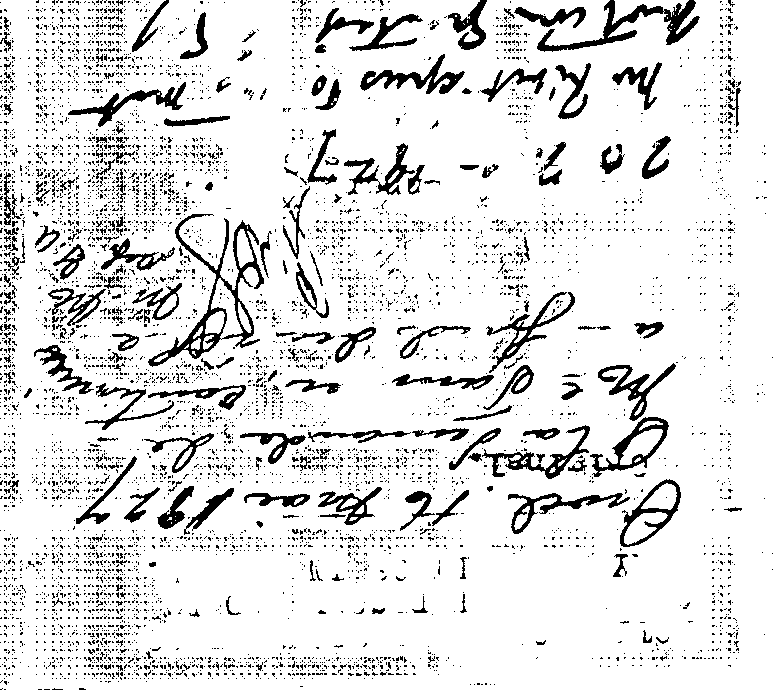} &
\includegraphics[height=2.40cm]{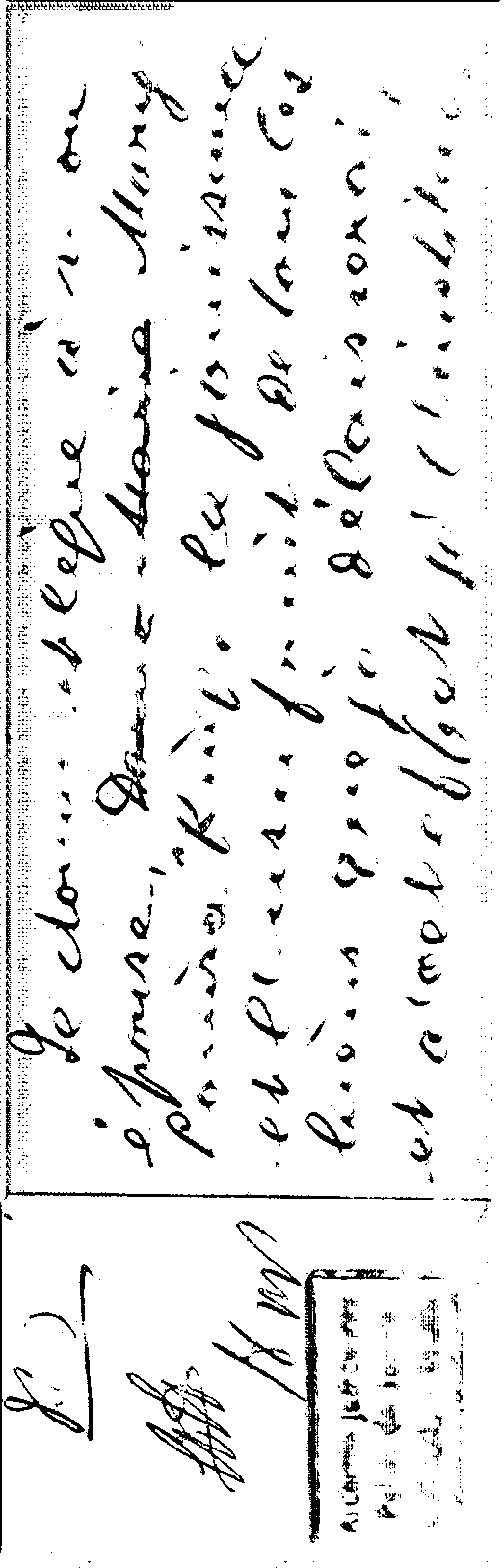}
\\

$F4+$NIR Edge-Guided &
\includegraphics[height=2.40cm]{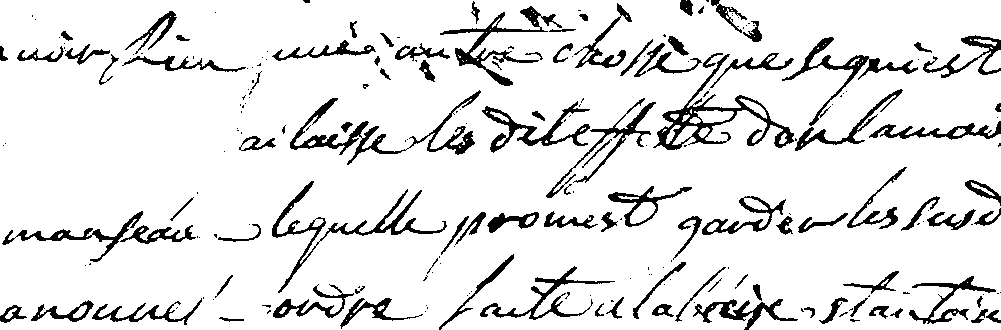} &
\includegraphics[height=2.40cm]{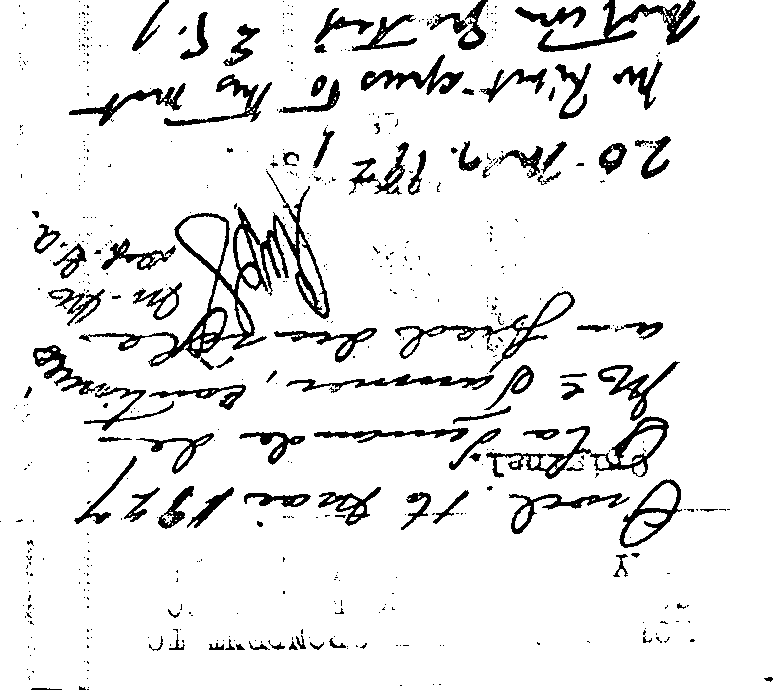} &
\includegraphics[height=2.40cm]{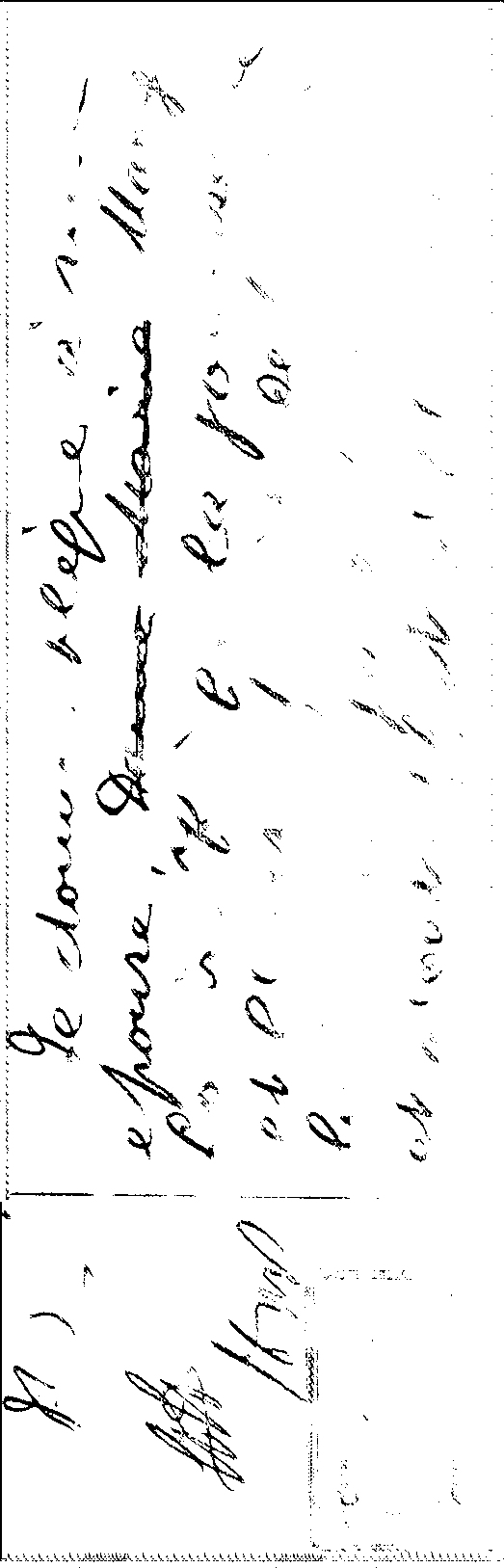}
\\

Ground Truth &
\includegraphics[height=2.40cm]{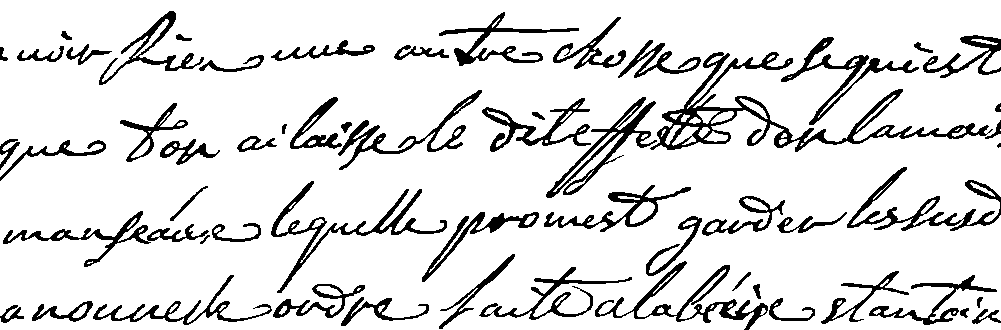} &
\includegraphics[height=2.40cm]{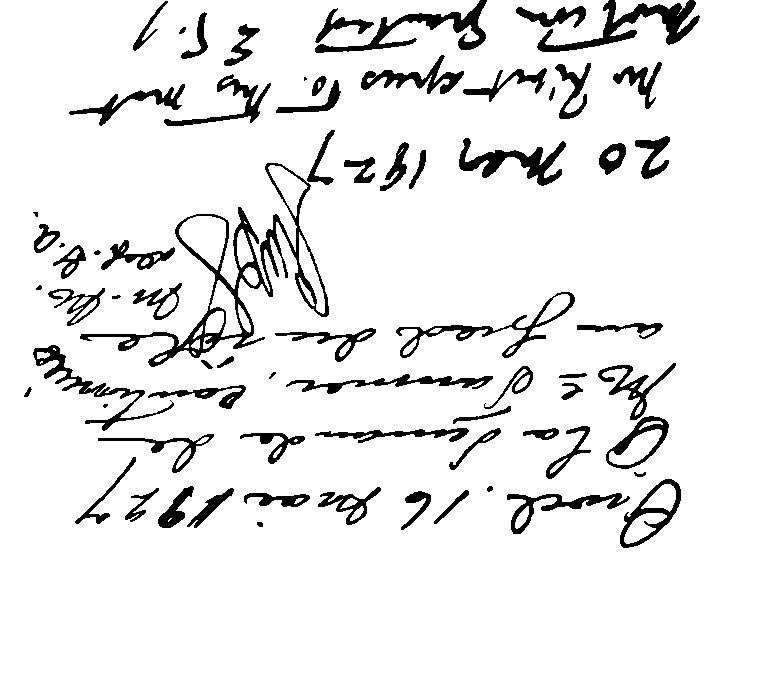} &
\includegraphics[height=2.40cm]{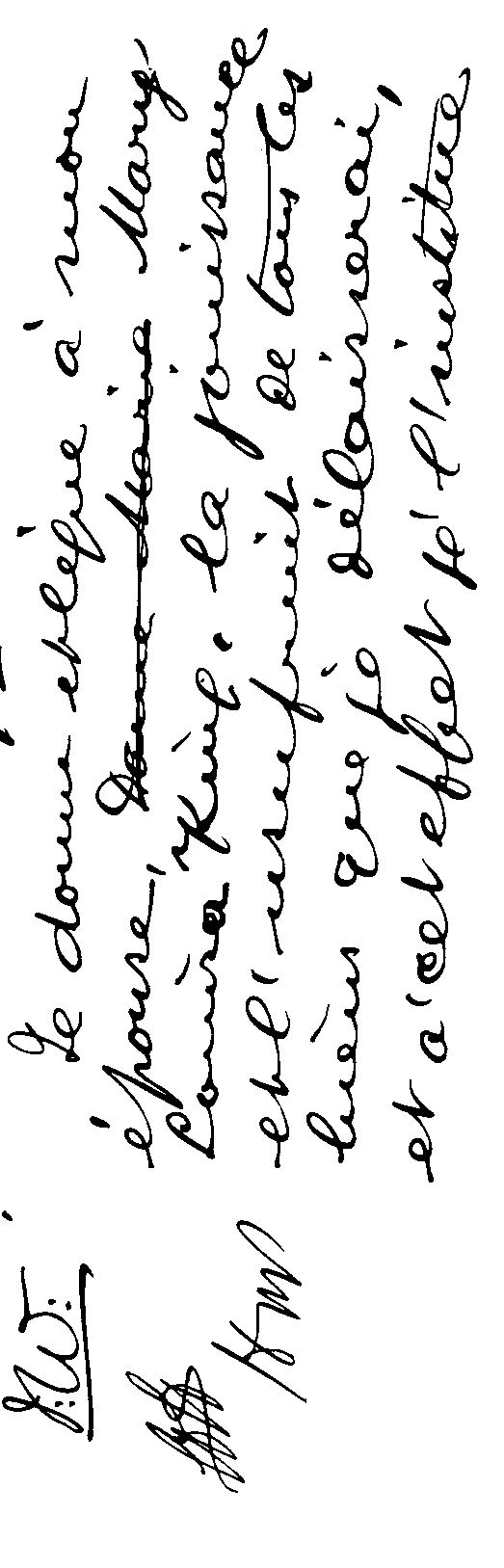}
\\

\bottomrule
\end{tabular}

\caption{Qualitative comparison of channel-attention and wavelength-specific edge-guided models ($F2$--$F4+\text{NIR}$). The proposed $F3+$NIR model is highlighted in bold.}
\label{fig:qualitative_b}
\end{figure*}

Figures~\ref{fig:qualitative_a} and~\ref{fig:qualitative_b} compare nine architectures on held-out documents $z27$, $z35$, and $z80$, covering bleed-through, illumination drop-offs, and faint strokes to assess foreground preservation, connectivity, and background suppression.

Mid AVG ($L=3$) recovers primary text but shows stroke fragmentation and boundary degradation because unweighted averaging treats text-bearing and noise-dominated channels equally. CBAM, SE, and Visible/NIR gating reduce background noise but introduce non-uniform artifacts.

The two four-band variants clarify the structural loss function's role. An edge-derived channel without auxiliary supervision ($\lambda_{\text{edge}}=0$) does not constrain feature representation space. Incorporating $\lambda_{\text{edge}}=5$ improves stroke continuity and suppresses isolated false positives, matching the quantitative gains observed in $\mathrm{F_1}$ and $\mathrm{DRD}$. 

Wavelength-specific guidance shows clear structural advantages. Among $F2+\text{NIR}$, $F3+\text{NIR}$, and $F4+\text{NIR}$, $F3+\text{NIR}$ gives stronger stroke reconstruction without boundary erosion or false-positive expansion. Deriving guidance from a targeted visible--NIR pair while retaining the full multispectral input preserves boundaries in severe degradation zones, supporting its $F_1$, p-FM, and IoU leads.

Residual errors persist where local background contrast approaches zero, aligning with metric-specific trades ($\text{e.g.}$, $F2+\text{NIR}$ obtaining minimal $\mathrm{NRM}$). Visual evidence nevertheless confirms that targeted multispectral structural guidance combined with adaptive intermediate fusion provides robust full-page binarization.
% Include challenging full-page examples and failure cases.

% ============================================================
% 6 LIMITATIONS
% ============================================================

\section{Limitations and Future Work}
\label{sec:limits}

Several scope and computational limitations warrant further investigation.

\textbf{Spectral Pair Selection and Dataset Scope:}
Evaluation uses 30 SMADI documents across four bands ($500$, $600$, $700$, and $1000~\mathrm{nm}$). Although five-fold document-level cross-validation ensures strict isolation, the dataset limits media, ink, and degradation diversity. Thus, $F3+\text{NIR}$ ($600/1000~\mathrm{nm}$) is not universally optimal; broader corpora and data-driven wavelength selection remain future work.

\textbf{Loss Formulation and Structural Guidance Filters:}
The edge-aware loss weight ($\lambda_{\mathrm{edge}}=5$) and Sobel edge extraction were evaluated under controlled settings rather than exhaustive search. Sobel filters also respond to high-gradient bleed-through, paper grain, and stains, occasionally preserving non-text artifacts. Future work could use learned text-aware descriptors, multiscale edges, or confidence-weighted losses to balance recovery and noise suppression.

\textbf{Computational Complexity and Full-Page Inference:}
The generator contains $56.31\times 10^6$ parameters and requires approximately $20.68~\mathrm{GMACs}$ per $256\times 256$ patch. GPU latency is feasible for offline processing, but 20\% reconstruction overlap increases page-level computation. Deployment-oriented work should explore branch sharing, distillation, lightweight fusion, and fully convolutional page inference.

Future work will therefore emphasize cross-dataset generalization, adaptive spatial-spectral guidance, and architectural compression.

\section{Conclusion}
\label{sec:conclusion}

This study established a targeted multispectral intermediate-fusion framework for historical-document binarization, showing that physically guided regulation improves on simple averaging and generic attention. At feature depth $L=3$, learned edge-guided fusion replaces uniform channel weighting. A dedicated $F3+\text{NIR}$ ($600/1000~\mathrm{nm}$) pair supplies structural guidance while the full $F2,F3,F4,F7$ observation remains available, preserving character boundaries while suppressing complex degradation.
Across 30 SMADI pages under five-fold document-level cross-validation, $F3+\text{NIR}$ achieved statistically significant gains in $\mathrm{F_1}$, $\mathrm{p\text{-}FM}$, and $\mathrm{IoU}$ over fixed averaging and generic CBAM/SE attention. Patch overlap and runtime analyses confirmed that a $20\%$ patch reconstruction overlap provides a practical deployment compromise for full-page GPU inference.
The results confirm that physical wavelength selection governs distinct error profiles. While targeted $F3+\text{NIR}$ guidance delivers optimal overall foreground recovery and structural continuity, alternative pairings favor NRM or DRD. Targeted structural guidance therefore effectively regulates intermediate multispectral features and motivates future research into learned dynamic spectral selection and parameter-efficient document-scale architectures.

% ============================================================
% DECLARATIONS
% ============================================================
\backmatter

\section*{Statements and Declarations}

\textbf{Funding.}
This study is independent research carried out under the University of Peradeniya Multidisciplinary Research Grant Scheme (MRG\_2025\_06).

\textbf{Competing interests.}
The authors declare no competing interests.

\textbf{Author contributions.}
Yasas Savinda: Conceptualization, methodology, software, validation,
formal analysis, investigation, visualization, and writing--original draft.
Maheshi Buddhinee Dissanayake: Conceptualization, supervision, methodology,
funding and grant management, formal analysis, and writing--review and editing.
Both authors read and approved the final manuscript.

\textbf{Data availability.}
The experiments use the SMADI multispectral historical-document dataset
associated with the ICDAR 2015 MultiSpectral Text Extraction benchmark
\cite{MSTEx2015}. The original SMADI dataset and benchmark information are
publicly available for non-commercial research from the
\href{https://tc11.cvc.uab.es/datasets/SMADI_1}
{official IAPR TC-11 SMADI repository}.
Information on the dataset organization and file structure required to reproduce
the experiments is available in the
\href{https://drive.google.com/drive/folders/1WzY3u1szIYvi4AnDOCd9l_UewnaL5TzZ}
{supplementary dataset information folder}.
The experimental outputs and analysis results generated in this study are
available in the
\href{https://drive.google.com/drive/folders/1lknwd1YuHB7NNDaUnZhFvW4GM11PLgTU}
{supplementary results repository}.

\textbf{Code availability.}
The training, evaluation, and analysis code supporting the findings of this
study is available in the
\href{https://drive.google.com/drive/folders/1WlrKyYTde5DPSoAZA571sgpkcaweldxA}
{supplementary code repository}.

\textbf{Ethics approval.}
Not applicable.

\textbf{Consent to participate.}
Not applicable.

\textbf{Consent for publication.}
Not applicable.

\bibliography{references}

\end{document}